\pdfoutput=1 
\documentclass{article} 
\usepackage[final]{colm2026_conference}

\usepackage{microtype}
\usepackage{lineno}

\usepackage{amsmath,amsfonts,bm}

\def\eqref#1{equation~\ref{#1}}

\def\1{\bm{1}}

\DeclareMathAlphabet{\mathsfit}{\encodingdefault}{\sfdefault}{m}{sl}
\SetMathAlphabet{\mathsfit}{bold}{\encodingdefault}{\sfdefault}{bx}{n}

\usepackage{hyperref}
\usepackage{url}
\usepackage{kotex}
\usepackage{amsmath,amssymb,amsfonts}
\usepackage{algorithm} 
\usepackage{algpseudocode}
\usepackage{graphicx}
\usepackage{adjustbox}
\usepackage{textcomp}
\usepackage{verbatim}
\usepackage{multirow}
\usepackage{diagbox}
\usepackage{booktabs}
\usepackage{tabularx}
\usepackage{comment}
\usepackage{siunitx}
\usepackage[most]{tcolorbox}
\usepackage{dashrule}
\usepackage{caption}
\usepackage{enumitem}
\usepackage{pifont}
\usepackage{makecell} 
\usepackage{wrapfig}
\usepackage[dvipsnames]{xcolor}

\hypersetup{
  colorlinks=true,
  linkcolor=blue,
  citecolor=blue,
  urlcolor=blue,
  pdftitle={Guard Vector: Beyond English LLM Guardrails with Task-Vector Composition and Streaming-Aware Prefix SFT},
  pdfauthor={Wonhyuk Lee, Youngchol Kim, Yunjin Park, Junhyung Moon, Dongyoung Jeong, Wanjin Park}
}

\title{Guard Vector: Beyond English LLM \\ Guardrails with Task-Vector Composition \\ and Streaming-Aware Prefix SFT}

\author{\parbox{\dimexpr\textwidth-2\tabcolsep\relax}{\centering
\vspace*{10pt}
\textbf{Wonhyuk Lee}\quad\textbf{Youngchol Kim}\quad\textbf{Yunjin Park}\\[8pt]
\textbf{Junhyung Moon}\quad\textbf{Dongyoung Jeong}\quad\textbf{Wanjin Park}\thanks{Corresponding author.}\\[10pt]
{\mdseries KT Corporation, Responsible AI Department}}}

\newcommand{\dashline}{\par\noindent\hdashrule{\linewidth}{0.6pt}{4pt 2pt}\par}
\newcommand{\cmark}{\textcolor{ForestGreen}{\ding{51}}} 
\newcommand{\xmark}{\textcolor{BrickRed}{\ding{55}}}    
\tcbset{
  sysbox/.style={
    enhanced,
    colback=white,
    colframe=black,
    boxrule=1pt,
    rounded corners,
    title filled,
    colbacktitle=black,
    coltitle=white,
    fonttitle=\bfseries,
    titlerule=0pt,
    toptitle=3mm,
    bottomtitle=3mm,
    lefttitle=3mm,
    righttitle=3mm,
    before skip=20pt,
    after skip=20pt,
  }
}

\begin{document}

\ifcolmsubmission
\linenumbers
\fi

\maketitle

\begin{abstract}
We introduce \emph{Guard Vector}, a safety task vector computed as the parameter difference between a guardrail model (Guard Model) and a same-architecture pretrained language model. Composing this vector with a target language model yields a \emph{Target Guard Model (TGM)}. We then adapt TGM with a streaming-aware approach that combines \emph{prefix-based training} and evaluation with a classifier that produces a \emph{single-token output}. With this composition alone, TGM improves classification quality over established Guard Models across standard safety suites and enables language extensibility to Chinese, Japanese, and Korean, requiring neither additional training nor target language labels for this composition step. It also demonstrates model portability across two widely used public guardrail backbones, Llama and Gemma. With prefix SFT (supervised fine-tuning), TGM preserves classification quality under streaming by aligning the behavior between prefix inputs and full-text inputs. The single-token output design increases throughput and reduces latency. Together, these components reduce data and compute requirements while promoting streaming-aware evaluation practices, thereby contributing to a more responsible AI ecosystem.

\begin{center}
\vspace{-0.5em}
\adjustbox{valign=m}{\includegraphics[height=1.5em]{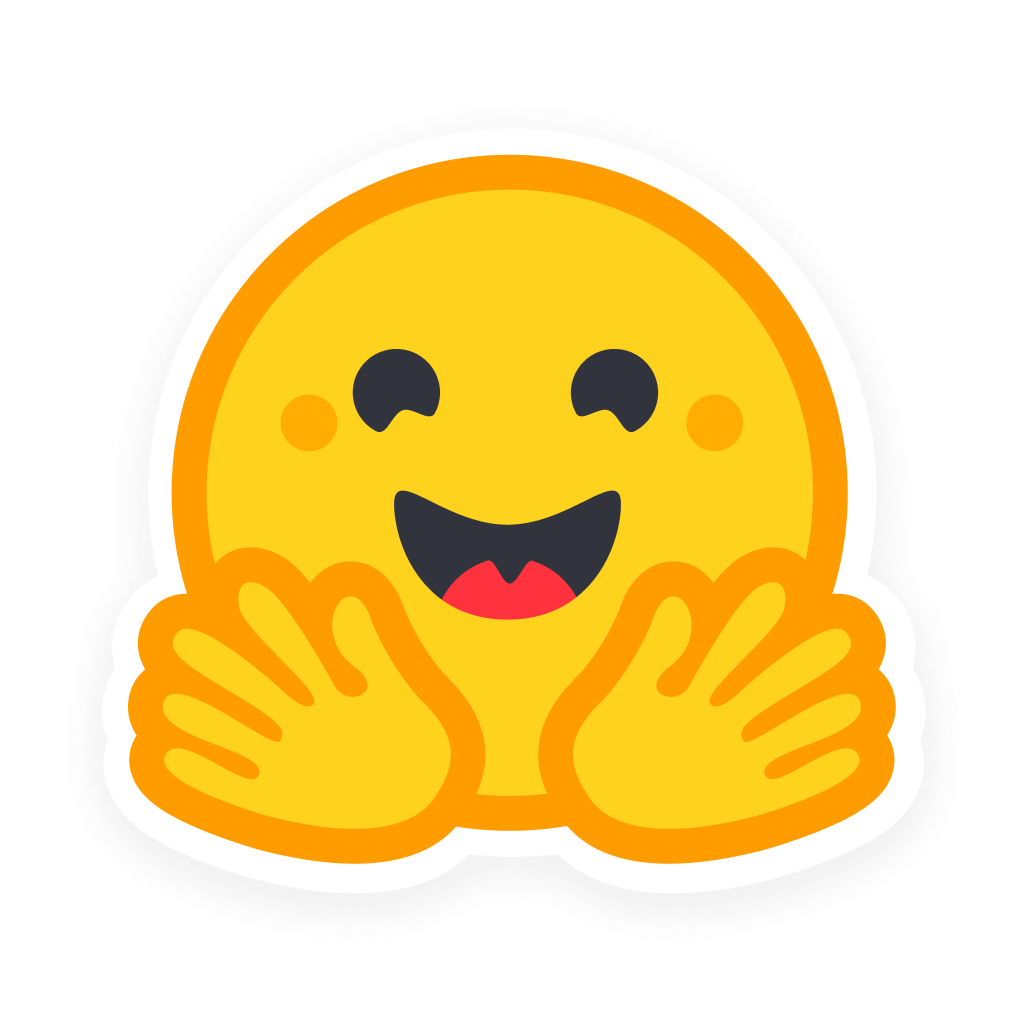}}\enspace
\href{https://huggingface.co/K-intelligence/Llama-SafetyGuard-Content-Binary}{\textbf{Model: K-intelligence/Llama-SafetyGuard-Content-Binary}}
\vspace{0.3em}
\end{center}
\end{abstract}

\section{Introduction}
Large language models (LLMs) are increasingly deployed across real-world applications, including online search engines, counseling, software development, finance, healthcare, and law \citep{mialon2023augmented,qu2025tool}. This widespread application has increased demand for safety, reflecting concerns about the risks of incorporating LLMs in these sensitive domains. The predominant approach to ensuring LLM-safety is safety alignment, yet it suffers from inherent limitations such as vulnerability to novel jailbreaks, limited coverage of unseen risk categories, and inconsistent performance in cross-lingual settings \citep{bai2022constitutional}. These shortcomings have motivated the use of guardrails as a complementary safeguard. Guardrails are specialized safety layers that classify or block model inputs and outputs based on predefined risk policies and have emerged as a practical mechanism for enforcing such safeguards \citep{openai2022moderation, ghosh-etal-2025-aegis2}.

However, implementing guardrails for non-English languages remains challenging due to the core reliance on English-centric models and policies \citep{dubey2024llama3herdmodels, team2024gemma}, the high cost of alignment pipelines that rely on supervised fine-tuning (SFT) and additional training \citep{zeng2024shieldgemma, wildguard2024}, and the limited availability of labeled datasets in target languages \citep{costa2022no}.
Another major challenge arises from streaming interactions, which are the default mode in many production environments. Although LLMs generate responses token by token in both offline and streaming modes, streaming additionally exposes partial outputs to users (and thus to guardrails) before the full response is complete, making it crucial for guardrails to provide immediate feedback and detect risk signals at early stages, especially for long outputs. Yet most guardrail research has focused on offline evaluation with access to full-text \citep{dubey2024llama3herdmodels,zeng2024shieldgemma}, with little systematic verification of standardized streaming metrics or parity with offline performance. As a result, ensuring accurate, high-throughput and low-latency guardrail decisions during streaming conditions remains an open challenge.

\begin{figure}[t]
\centerline{\includegraphics[width=1\linewidth]{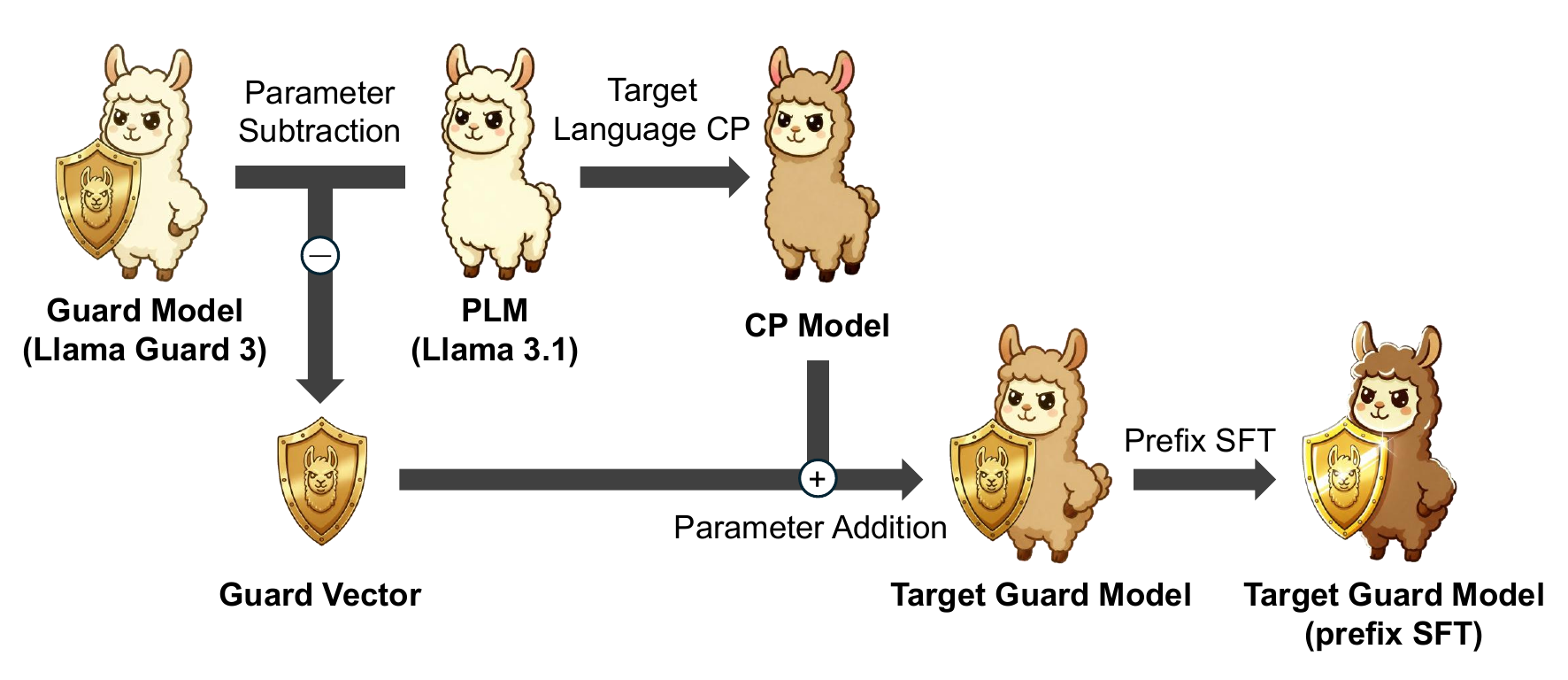}}
\caption{Pipeline of task-vector composition and streaming adaptation. 
A \emph{Guard Vector} is computed as the parameter difference between a Guard Model and a pretrained language model (PLM) of the same architecture. 
This vector is composed with a continual pretraining model (CP Model) in the target language to yield the Target Guard Model (TGM). 
Finally, streaming-aware prefix SFT with a single-token classifier aligns prefix and full-text behavior.}
\label{fig:main}
\end{figure}

To address these limitations in non-English guardrail development, we propose \textbf{Guard Vector}, a \textbf{task-vector composition} method that transfers safety behaviors to target language models using publicly available weights and requiring neither additional training nor target language labels. Guard Vector is a practical instantiation of task-vector arithmetic \citep{ilharco2022editing} tailored to multilingual safety transfer, rather than a new general-purpose composition rule; our contribution is the deployment-oriented combination of label-free safety transfer with a streaming-aware guardrail protocol. Specifically, a \emph{Guard Model} is a pretrained LLM that is fine-tuned for safety classification (i.e., guardrail) while a \emph{pretrained language model (PLM)} is the same-architecture model without safety fine-tuning. The parameter difference between the Guard Model and PLM yields a \emph{Guard Vector}. We then compose this resulting Guard Vector with a \emph{continual pretraining model (CP Model)}, a same-architecture model further pretrained on large-scale non-English corpora, to obtain a \emph{Target Guard Model (TGM)} with safety behaviors transferred to the target language.
To address streaming evaluation and efficiency, we further introduce Target Guard Model with \textbf{streaming-aware prefix SFT (TGM (prefix SFT))}, a variant of TGM fine-tuned with cumulative prefixes and a single-token output classifier.
This adaptation enables early detection from partial inputs while maintaining parity with offline evaluation and reducing latency through single-step classification. 
Figure~\ref{fig:main} summarizes the complete pipeline from a given Guard Model, PLM, Guard Vector, CP Model, TGM, to TGM (prefix SFT).

We first compare the baseline Guard Model and TGM under the same architecture and observe that composition alone improves classification quality compared to the baseline. 
Next, we demonstrate portability of the method across both Llama and Gemma architectures using the same composition procedure. 
With the addition of prefix SFT, TGM further surpasses both the baseline’s prefix SFT variant and the baseline Guard Model. 
Finally, we evaluate TGM under both offline (full-text) and streaming (prefix) regimes, reporting standardized streaming metrics and demonstrating parity with offline evaluation.

\paragraph{The contributions of this paper are summarized as follows:}
\begin{itemize}[leftmargin=*]

\item \textbf{Addressing two practical gaps in guardrail deployment.}
We show that (a) safety behaviors can be transferred to target language models \emph{without additional training or target language labels} via composition of public weights, and (b) streaming guardrails can be trained and evaluated for \emph{parity with offline} through a standardized prefix-based protocol with a single-token classifier. The training-free, label-free property applies to the Guard Vector composition step; prefix SFT is a separate supervised streaming-specialization stage that uses curated prefix data.

\item \textbf{Guard Vector: task-vector composition from public weights.}
We define the Guard Vector as the parameter difference between a Guard Model and a same-architecture PLM, and compose it with a target language CP Model to obtain a TGM — requiring no additional training or labels. This enables portability across Llama and Gemma families and extensibility to Chinese, Japanese, and Korean.

\item \textbf{Streaming-aware prefix SFT for low-latency guardrails.}
We propose a prefix-based training and evaluation recipe where prefix labels are derived from full-response labels and unsafe-onset localization, decisions are applied to cumulative prefixes, and inferences are used as a single-token classifier. TGM (prefix SFT) achieves offline–streaming parity while improving throughput and latency.

\item \textbf{Responsible AI impact.}
Our pipeline provides a training-light path to reliable non-English guardrails from public weights, reducing data and compute burden while promoting streaming-aware evaluation practices.

\end{itemize}

\section{Related Work}
\label{sec:related_work}
\begin{minipage}[t]{0.60\columnwidth}
\vspace{0pt}

\paragraph{Guardrail Models}
Open-weight guardrails generally follow a \emph{backbone + supervised fine-tuning (SFT) or low-rank adaptation (LoRA)} recipe on English-centric data. Llama Guard 3 and 4 apply SFT on Llama backbones, with language coverage centered on English and only limited non-English reporting \citep{dubey2024llama3herdmodels,meta_llama_guard4_2025}. NeMoGuard (AEGIS 2.0) releases LoRA adapters on Llama-3.1-Instruct; despite broad safety categories, its dataset primarily targets English, limiting applicability in non-English contexts \citep{ghosh-etal-2025-aegis2, dubey2024llama3herdmodels}. ShieldGemma is built on Gemma-2, defines the problem in English and aggregates per-category judgments at inference, which can add latency relative to single-pass binary classification \citep{zeng2024shieldgemma,team2024gemma}. A compact landscape is summarized in Table~\ref{tab:guardrail-landscape}.

\paragraph{Further related work.}
See Appendix~\ref{app:related} for task-vector arithmetic, streaming-aware guardrails, and evaluation-protocol details.

\end{minipage}\hfill
\begin{minipage}[t]{0.36\columnwidth} 
\vspace{0pt}
\captionsetup{type=table}
\footnotesize
\setlength{\tabcolsep}{3pt}
\renewcommand{\arraystretch}{1.05}
\begin{tabular}{@{}lccc@{}} 
\toprule
\textbf{Model} & \textbf{BB} & \textbf{NT} & \textbf{CJK} \\
\midrule
Llama Guard 3   & L   & \xmark & \xmark \\
Llama Guard 4   & L   & \xmark & \xmark \\
NeMoGuard       & L   & \xmark & \xmark \\
ShieldGemma     & G   & \xmark & \xmark \\
\cmidrule(lr){1-4}
Target Guard Model      & L/G & \cmark & \cmark \\
\bottomrule
\end{tabular}
\caption{Comparison across backbone (BB; L{=}Llama, G{=}Gemma), need for additional training (NT; \cmark~means none), and Chinese/Japanese/Korean coverage (CJK). Target Guard Model spans both backbones and is the only entry in this comparison that combines training-free composition with evaluated Chinese/Japanese/Korean coverage.}
\label{tab:guardrail-landscape}
\end{minipage}

\section{Methodology}
\label{sec:method}

We compute a safety task vector (Guard Vector) as the parameter difference between a guardrail model (Guard Model) and a pretrained language model (PLM) of the same architecture. Given a target language continual pretraining model (CP Model) with the same architecture, we compose the Guard Vector with the CP model to obtain a Target Guard Model (TGM). To support streaming regimes, we introduce a streaming-aware prefix SFT recipe that supervises decisions on cumulative prefixes and uses a single-token output classifier to align prefix and full-text behavior while improving efficiency.

\subsection{Guard Vector Composition}
\label{subsec:setup-compose}

\paragraph{Notation and assumptions.}
Assume all models share the same architecture.
Let $\theta_{\mathrm{PLM}}, \theta_{\mathrm{GM}}, \theta_{\mathrm{CP}}$ denote the parameters of PLM, GM, and the target language CP Model, respectively.
Each parameter set $\theta$ is a map from parameter names to tensors; write $\mathrm{keys}(\theta)$ for its key set and index tensors by $t$ (e.g., $\theta[t]$).
We define the excluded parameter types and the composition domain as
\begin{equation}
\label{eq:scope}
\mathcal{E}=\{\text{embeddings},\,\text{lm\_head},\,\text{LayerNorm}\},\;
S=\big(\mathrm{keys}(\theta_{\mathrm{PLM}})\cap \mathrm{keys}(\theta_{\mathrm{GM}})\cap \mathrm{keys}(\theta_{\mathrm{CP}})\big)\setminus \mathcal{E}.
\end{equation}
with an additional requirement of exact tensor-shape equality; if shapes mismatch for a key, that key is not included in $S$.
We exclude $\mathcal{E}$ (token embeddings, the LM head, and LayerNorm) primarily for cross-checkpoint compatibility and stability rather than as a performance optimization: token embeddings and the LM head are tied to each checkpoint's vocabulary and output space and can become incompatible when vocabularies or added label tokens differ, and LayerNorm parameters are known to be sensitive under task-vector composition \citep{xiong2020onlayernorm,shirafuji2024bias}; Appendix~\ref{app:compose-ablation} ablates the scaling factor, the excluded parameter groups, and the layer range.
We write $\theta_{\mathrm{TGM}}$ for the composed model and $\theta_{\mathrm{TGM\text{-}SFT}}$ for its prefix-SFT adaptation.
Labels are binary, $y\in\{\text{SAFE},\text{UNSAFE}\}$, mapped to $\{0,1\}$ with UNSAFE$=1$.

\paragraph{Guard Vector and composition rules.}
For each $t\in S$, define the Guard Vector as the parameter difference
\begin{equation}
\label{eq:gv}
V_{\mathrm{GV}}[t] \;=\; \theta_{\mathrm{GM}}[t] \;-\; \theta_{\mathrm{PLM}}[t].
\end{equation}
Obtain the \textbf{Target Guard Model (TGM)} by adding the Guard Vector to the CP Model:
\begin{equation}
\label{eq:tg}
\theta_{\mathrm{TGM}}[t] \;=\; \theta_{\mathrm{CP}}[t] \;+\; V_{\mathrm{GV}}[t], \quad t\in S; \qquad
\theta_{\mathrm{TGM}}[t] \;=\; \theta_{\mathrm{CP}}[t], \quad t\notin S.
\end{equation}
No scaling factor is applied. A summary of the procedure appears in Algorithm~\ref{alg:guard-vector-compose}.

\begin{algorithm}[t]
\caption{Compose Target Guard Model with Guard Vector}
\label{alg:guard-vector-compose}
\begin{algorithmic}[1]
\Require $\theta_{\mathrm{PLM}}$ (e.g., Llama~3.1), $\theta_{\mathrm{GM}}$ (e.g., Llama Guard~3), $\theta_{\mathrm{CP}}$ (target language CP Model); same architecture
\Ensure $\theta_{\mathrm{TGM}}$
\State $\mathcal{E}\leftarrow\{$embeddings, lm\_head, LayerNorm parameters$\}$
\State $S \leftarrow \big(\mathrm{keys}(\theta_{\mathrm{PLM}})\cap\mathrm{keys}(\theta_{\mathrm{GM}})\cap\mathrm{keys}(\theta_{\mathrm{CP}})\big)\setminus\mathcal{E}$
\For{each $t\in S$}
  \State $V_{\mathrm{GV}}[t]\leftarrow \theta_{\mathrm{GM}}[t]-\theta_{\mathrm{PLM}}[t]$
  \State $\theta_{\mathrm{TGM}}[t]\leftarrow \theta_{\mathrm{CP}}[t]+V_{\mathrm{GV}}[t]$
\EndFor
\For{each $t\notin S$}
  \State $\theta_{\mathrm{TGM}}[t]\leftarrow \theta_{\mathrm{CP}}[t]$
\EndFor
\State \Return $\theta_{\mathrm{TGM}}$
\end{algorithmic}
\end{algorithm}

\subsection{Target Guard Model (prefix SFT): Streaming-Aware Prefix Training and Latency-Aware Design}
\label{subsec:tg-sft}

\paragraph{Streaming-aware prefix training.}
We propose \textbf{prefix SFT} for $\theta_{\mathrm{TGM}}$ to support streaming deployment.
For a model-generated response $r$ of length $L$, we construct cumulative prefixes under a monotone prefix schedule $\mathcal{K}(r)\subseteq\{1,\ldots,L\}$:
\begin{equation}
\label{eq:prefix-chunks}
\mathcal{C}(r) \;=\; \{\, r_{1:K} \mid K \in \mathcal{K}(r) \,\}.
\end{equation}
This schedule can be defined at different granularities (characters, tokens, or sentence boundaries).
Unless otherwise noted, we instantiate a character-based schedule $\mathcal{K}(r)=\{100,200,\ldots,L\}$:
it is tokenizer-agnostic (avoids mismatches between the generator and the guardrail), provides predictable periodic checking for long outputs, and reduces bookkeeping and context-tracking overhead.
The fixed 100-character step is an implementation choice; because scripts differ in information density, an identical character count need not correspond to identical semantic granularity across languages, so any cross-language comparison based solely on character count should be interpreted with caution.

Prefix labels are derived from the full-response label and, for UNSAFE responses, the harmful-onset location.
If the original response is SAFE, all prefixes are labeled SAFE; if the original response is UNSAFE, prefixes strictly before the first occurrence of harmful content are labeled SAFE, and prefixes at or after that point are labeled UNSAFE.
We discard sequences that violate monotonicity (e.g., SAFE$\to$UNSAFE$\to$SAFE) and UNSAFE cases with no detected harmful prefix.
This supervision targets early detection under streaming regimes.

\paragraph{Prefix-level class balancing.}
The $\mathcal{C}(\cdot)$ expansion alters class balance at the prefix level: when UNSAFE responses tend to be longer, more prefixes are labeled UNSAFE, while shorter SAFE responses contribute fewer prefixes. To prevent this drift from biasing training, we rebalance the prefix pool to a 1:1 SAFE/UNSAFE ratio by downsampling the majority class at the prefix level. We considered class-weighted losses as an alternative, but found simple resampling sufficient in our setting. Detailed training dataset counts and resampling settings are provided in Appendix~\ref{app:training}.

\paragraph{Single-token classification.}
We classify with a single-token output: concretely, we use the model’s next-token distribution restricted to two reserved label tokens
$\mathcal{V}_y=\{v_{\texttt{<SAFE>}},\,v_{\texttt{<UNSAFE>}}\}$ added to the tokenizer.
Given a prefix prompt $p(r_{1:K})$, let $z_{\mathrm{SAFE}}$ and $z_{\mathrm{UNSAFE}}$ denote the next-token logits for these two labels, and define
\begin{equation}
\label{eq:prob-unsafe}
p_\theta(\mathrm{UNSAFE}\mid p(r_{1:K})) \;=\;
\frac{\exp(z_{\mathrm{UNSAFE}})}{\exp(z_{\mathrm{SAFE}})+\exp(z_{\mathrm{UNSAFE}})}.
\end{equation}
We train with binary cross-entropy (UNSAFE$=1$):
\begin{equation}
\label{eq:bce}
\mathcal{L}(\theta)
= - \sum_{(r_{1:K},\,y)}
\Big[ y \,\log p_\theta(\mathrm{UNSAFE}\mid p(r_{1:K}))
      + (1-y)\,\log\big(1-p_\theta(\mathrm{UNSAFE}\mid p(r_{1:K}))\big) \Big].
\end{equation}
At inference, we predict using the unsafe classification threshold $\tau$ (see §\ref{subsec:protocol}):
\begin{equation}
\label{eq:decision-infer}
\hat{y}\;=\;\mathbf{1}\!\big[p_\theta(\mathrm{UNSAFE}\mid p(r_{1:K}))\ge \tau\big].
\end{equation}
This single-token output uses a single forward pass per prefix (no multi-token decoding) and reduces latency compared to generation-based evaluators that emit multi-token rationales.

\paragraph{Summary.}
Prefix SFT supervises the model on monotone prefix inputs to induce early detection.
The single-token classification objective enables low-latency, single-forward-pass inference.
Combined in $\theta_{\mathrm{TGM\text{-}SFT}}$, these components align with response streaming regimes. §\ref{sec:results-exp1}, \ref{sec:results-exp2} demonstrate improvements in classification quality and runtime efficiency, respectively.

\section{Experimental Setup} 
\label{sec:exp-setup}

\subsection{Datasets}
\label{subsec:datasets}
We evaluate three Korean datasets: a proprietary Harmlessness Evaluation Dataset, the public Kor Ethical QA \citep{kor_ethical_qna2024}, and a proprietary Helpfulness Evaluation Dataset. The Harmlessness Evaluation Dataset and Kor Ethical QA are evaluation-only; they are never used for training and are used in their entirety without splitting or sampling. The Helpfulness Evaluation Dataset contains only SAFE responses. 
Due to its limited sample count, we partition this over-refusal corpus into a training split and a held-out evaluation split (the held-out split is the Helpfulness Evaluation Dataset); 
on this dataset we report accuracy since the positive class (UNSAFE) is absent (see §\ref{subsec:protocol}). Full details of the AI risk taxonomy, dataset construction, and statistics are in Appendices~\ref{app:risk-taxonomy} and~\ref{app:evaluation-dataset}.

\subsection{Models}
\label{subsec:models}
Comparison targets are as follows:
\begin{itemize}[leftmargin=*]
\item \textbf{LG3}: Llama Guard 3 \citep{dubey2024llama3herdmodels}.
\item \textbf{Kanana Safeguard}: Korean guardrail baseline \citep{KananaSafeguard}.
\item \textbf{LG3 (prefix SFT)}: LG3 with 100-character cumulative prefix SFT applied.
\item \textbf{TGM}: Target Guard Model. Guard Vector composed with a language-specific CP Model (§\ref{sec:method}).
\item \textbf{TGM (full-text SFT)}: TGM with full-text SFT applied.
\item \textbf{TGM (prefix SFT)}: TGM with 100-character cumulative prefix SFT applied.
\end{itemize}

\paragraph{Model setup note.}
Unless otherwise noted, all models are 8B-parameter variants. Except for §\ref{sec:results-exp3}, we use the Korean CP Model of \citet{varco2024llama} to build the TGM family.
Details of decoding and label mapping are specified in §\ref{subsec:protocol}.

\subsection{Evaluation Setup}
\label{subsec:protocol}

\paragraph{Task and data.}
Guardrails classify model responses (not user prompts). The positive class is UNSAFE.

\paragraph{Regimes.}
We evaluate under two regimes, offline (full-text) and streaming (prefix).
Offline uses the entire model response $r$ as input to the classifier.
Streaming mimics real-time display: the decision rule is applied to \emph{cumulative} character-level prefixes $r_{1:K}$ of the same response.
Unless otherwise noted, we adopt a character-based monotone schedule with base step $K=100$ characters (i.e., $K\in\{100,200,\ldots,|r|\}$), and all reported results follow this setup.

\paragraph{Decision pipelines.}
We use two inference pipelines depending on the model interface. Both are evaluated in offline (full-text) and streaming (prefix) regimes.

\begin{itemize}[leftmargin=*]
\item \textbf{Single-token output (SFT family).}
Inference follows §\ref{subsec:tg-sft}: we compute the unsafe probability from the two label-token logits (Eq.~\ref{eq:prob-unsafe}) and apply the unsafe classification threshold (Eq.~\ref{eq:decision-infer}). Unless otherwise noted, $\tau=0.5$ (this threshold applies only to this pipeline).
Under streaming, prefixes $r_{1:K}$ are evaluated in increasing $K$. Once any prefix is classified UNSAFE, we early-terminate for that instance and classify the instance as UNSAFE.
If no prefix is classified UNSAFE, the instance is SAFE at stream end.

\item \textbf{Generation-and-parse models (e.g., LG3, TGM without SFT).}
Decoding is deterministic for all models (temperature $=0$).
We parse each model’s textual judgment using its recommended schema and map it to a binary label.
Under streaming, the parser is run on each prefix and we early-terminate at the first prefix classified UNSAFE; otherwise the instance is SAFE at stream end.
\end{itemize}

\paragraph{Evaluation Metrics.}
Unless noted otherwise, all classification-quality metrics (F1, BER, Accuracy, TTD) are \emph{computed as rates in $[0,1]$ and reported in percentage units (0–100)}.

\begin{itemize}[leftmargin=*]
  \item \textbf{Classification quality.}
  \textbf{F1} is the binary F1 score with UNSAFE as the positive class (the harmonic mean of UNSAFE precision and recall; higher values indicate higher classification quality).
  \textbf{Balanced Error Rate (BER)} is $\tfrac{1}{2}\bigl(\mathrm{FPR}+\mathrm{FNR}\bigr)$ (lower BER value indicates higher classification quality),
  where False Positive Rate (FPR) misclassifies SAFE inputs as UNSAFE (\emph{over-refusal}) and
  False Negative Rate (FNR) misclassifies UNSAFE inputs as SAFE (\emph{missed risk}).
  BER captures the trade-off between these two error types.

  \item \textbf{All-SAFE datasets.}
  When a dataset contains only SAFE samples, F1 and BER are not informative because the positive class UNSAFE is absent.
  Therefore, we use \textbf{Accuracy} (higher Accuracy value indicates higher classification quality), defined as
  $\mathrm{Accuracy} = \tfrac{TP + TN}{N} = \tfrac{TN}{N} = 1 - \mathrm{FPR}$ since $N = TN + FP$.

  \item \textbf{Streaming-specific.}
  \textbf{Time to Detect (TTD)} for each UNSAFE sample is defined as $\mathrm{TTD} = \tfrac{\text{prefix length at first threshold crossing}}{\text{total response length}}$.
  We report mean TTD over detected UNSAFE cases; non-detections are excluded from this mean and are reflected by $\mathrm{FNR}$. TTD should therefore be interpreted together with FNR, since averaging only over detected cases can understate detection latency when FNR is non-zero.

  \item \textbf{Efficiency.} We assess throughput and latency using Queries per Second (QPS) and Tokens per Second (TPS; higher TPS value indicates better efficiency) and average latency per request (lower value indicates better efficiency).
\end{itemize}

\paragraph{System Prompts.}
System prompts use minimal instructions for SFT models and Kanana Safeguard, while LG3 and TGM follow LG3's default template.
Performance of TGM (prefix SFT) as a function of system prompt is summarized in Appendix~\ref{app:sys-prompt}.

Complete tables for precision, recall, FPR, and FNR are provided in Appendix~\ref{app:detail-t23}.

\section{Experimental Results}
\label{sec:exp-result}

\subsection{Results: Offline and Streaming Classification Quality (Experiment 1)}
\label{sec:results-exp1}

This section quantifies SAFE/UNSAFE classification quality on Korean evaluation datasets (§\ref{subsec:datasets}, Appendix~\ref{app:eval-data})
under offline (full-text) and streaming (prefix, $K{=}100$ characters) regimes, following the same evaluation setup (§\ref{subsec:protocol}). The summary metrics are \textbf{F1} and \textbf{BER}, and in streaming we additionally report \textbf{TTD}, 
the proportion of characters at which the first unsafe prediction occurs. Overall results are presented in Tables~\ref{tab:int-off-vs-stream}, \ref{tab:kor-ethical-off-vs-stream}, and detailed precision, recall, FPR, 
and FNR are reported in Appendix~\ref{app:detail-t23}.

\paragraph{Significant improvement over LG3 with Guard Vector composition alone.}
In offline, TGM showed consistent F1 increases compared to LG3: 
Harmlessness Evaluation Dataset showed \textbf{+9.57pp}, 
Kor Ethical QA showed \textbf{+11.51pp}. 
The same increases were maintained in streaming:
Harmlessness Evaluation Dataset \textbf{+6.88pp}, 
Kor Ethical QA \textbf{+8.27pp}.
BER likewise decreases relative to LG3 in both datasets and both regimes, mirroring the F1 gains.
These results show that Guard Vector composition alone transfers safety behaviors to target language models, requiring neither additional training nor target language labels. Consistent improvements across offline and streaming further support robustness and practical applicability. Appendix~\ref{app:no-gv-control} isolates this effect with a no-Guard-Vector control across Korean, Chinese, and Japanese.

\paragraph{TGM superior to LG3 in prefix SFT.}
Across both datasets and in both regimes, TGM (prefix SFT) attains higher F1 than LG3 (prefix SFT).
Offline: Harmlessness Evaluation Dataset \textbf{+2.07pp}, Kor Ethical QA \textbf{+3.59pp}.
Streaming: Harmlessness Evaluation Dataset \textbf{+1.85pp}, Kor Ethical QA \textbf{+3.00pp}.
These results indicate that applying prefix SFT to TGM—obtained by composing a Guard Vector with the CP Model—yields greater gains than applying prefix SFT directly to LG3.

\begin{table}[t] \centering \small \setlength{\tabcolsep}{4pt} \begin{tabular}{lrrrrrr} \toprule \textbf{Model} & \textbf{F1(off)} & \textbf{F1(str)} & \textbf{ΔF1} & \textbf{BER(off)} & \textbf{BER(str)} & \textbf{TTD(str)} \\ \midrule Llama Guard 3 & 82.05 & 85.64 & +3.59 & 15.23 & 12.63 & 49.60\% \\ Kanana Safeguard & 93.45 & 90.38 & -3.07 & 6.27 & 9.92 & 45.30\% \\ Llama Guard 3 (prefix SFT) & 96.31 & 96.51 & +0.20 & 3.58 & 3.42 & 53.40\% \\ \textbf{Target Guard Model} & 91.62 & 92.52 & +0.90 & 7.76 & 7.14 & 47.50\% \\ Target Guard Model (full-text SFT) & 98.84 & 83.61 & -15.23 & 1.16 & 19.57 & 40.80\% \\ \textbf{Target Guard Model (prefix SFT)} & \textbf{98.38} & \textbf{98.36} & -0.02 & \textbf{1.61} & \textbf{1.63} & 49.30\% \\ \bottomrule \end{tabular}
\caption{Harmlessness Evaluation Dataset: offline and streaming classification quality (prefix $K{=}100$; $\tau{=}0.5$; positive class = UNSAFE). $\Delta$F1 denotes streaming $-$ offline. Bold marks the best value per metric column among deployment-ready models (highest F1, lowest BER); the full-text SFT variant is a negative control excluded from this comparison, and $\Delta$F1 and TTD are secondary diagnostics that are not bolded.}
 \label{tab:int-off-vs-stream}
 \end{table}

\begin{table}[t] \centering \small \setlength{\tabcolsep}{4pt} \begin{tabular}{lrrrrrr} \toprule \textbf{Model} & \textbf{F1(off)} & \textbf{F1(str)} & \textbf{ΔF1} & \textbf{BER(off)} & \textbf{BER(str)} & \textbf{TTD(str)} \\ \midrule Llama Guard 3 & 83.29 & 86.45 & +3.16 & 14.32 & 12.16 & 58.60 \\ Kanana Safeguard & 80.20 & 73.94 & -6.26 & 24.46 & 35.08 & 51.10 \\ Llama Guard 3 (prefix SFT) & 94.16 & 94.79 & +0.63 & 5.52 & 4.96 & 62.70 \\ \textbf{Target Guard Model} & 94.80 & 94.72 & -0.08 & 4.96 & 5.25 & 54.30 \\ Target Guard Model (full-text SFT) & 98.19 & 71.54 & -26.65 & \,1.83 & 39.77 & 48.80 \\ \textbf{Target Guard Model (prefix SFT)} & \textbf{97.75} & \textbf{97.79} & +0.04 & \textbf{2.21} & \textbf{2.18} & 56.60 \\ \bottomrule \end{tabular}
\caption{Kor Ethical QA:  Offline and Streaming classification quality. Same setup as Table~\ref{tab:int-off-vs-stream}; see §\ref{subsec:protocol}. Bolding follows Table~\ref{tab:int-off-vs-stream}.}
\label{tab:kor-ethical-off-vs-stream}
\end{table}

\paragraph{Robust Korean Guardrail Performance.}
Across both regimes and both datasets, TGM (prefix SFT) shows higher F1 and lower BER than the Korean baseline Kanana Safeguard.
Offline: Harmlessness Evaluation Dataset F1 +4.93pp, BER -4.66pp; Kor Ethical QA F1 +17.55pp, BER -22.25pp.
Streaming: Harmlessness Evaluation Dataset F1 +7.98pp, BER -8.29pp; Kor Ethical QA F1 \textbf{+23.85pp}, BER \textbf{-32.90pp}.
These consistent gains indicate that TGM (prefix SFT) provides strong performance relative to existing baselines in our Korean evaluation settings.

\paragraph{Streaming parity: maintaining offline classification quality.}
TGM (prefix SFT) shows near-zero $\Delta$F1 (stream $-$ offline): \textbf{-0.02pp} on the Harmlessness Evaluation Dataset and \textbf{+0.04pp} on Kor Ethical QA.
BER changes are likewise minimal.
Thus, streaming classification quality matches offline behavior.
Using a shorter prefix step ($K=50$) yields the same parity (Appendix~\ref{app:abl-prefix}).

\paragraph{Streaming-aware training is necessary: full-text SFT degrades under streaming.}
TGM (full-text SFT) attains strong offline scores but degrades in streaming:
on the Harmlessness Evaluation Dataset, F1 drops by 15.23pp and BER worsens by 18.41pp;
on Kor Ethical QA, F1 drops by 26.65pp and BER worsens by 37.94pp.
These results underscore the need for prefix-based training to preserve early-decision quality under streaming.

\paragraph{Detection speed.}
In streaming, TTD was generally distributed in the 40--60\% range. Harmlessness Evaluation Dataset showed 40.8--53.4\%, and Kor Ethical QA showed 48.8--62.7\%. This suggests that many risks are detected before the full response is complete; these values should be interpreted together with FNR and the TTD-all analysis (Appendix~\ref{app:detail-t23}), and they support practical applicability along with the throughput (QPS, TPS) and average-latency results in §\ref{sec:results-exp2}.

\sisetup{
  group-separator       = {,},
  group-minimum-digits  = 4,
  input-ignore          = {,},
  output-decimal-marker = .
}

\begin{table}[t]
\centering
\scriptsize
\resizebox{\linewidth}{!}{%
\begin{tabular}{l rrr rrr rrr}
\toprule
\multirow{2}{*}{\textbf{Model}}
& \multicolumn{3}{c}{\textbf{QPS} $\uparrow$}
& \multicolumn{3}{c}{\textbf{TPS} $\uparrow$}
& \multicolumn{3}{c}{\textbf{Avg Latency (ms)} $\downarrow$} \\
\cmidrule(lr){2-4}\cmidrule(lr){5-7}\cmidrule(lr){8-10}
& {@200} & {@100} & {@10}
& {@200} & {@100} & {@10}
& {@200} & {@100} & {@10} \\
\midrule
Llama Guard 3 (LG3)
& 51.14 & 49.97 & 41.53
& 25,177 & 25,177 & 20,924
& 19.55 & 20.01 & 24.08 \\
\textbf{TGM (prefix SFT)}
& \textbf{77.50} & \textbf{77.49} & \textbf{83.42}
& \textbf{25,970} & \textbf{25,963} & \textbf{27,950}
& \textbf{12.90} & \textbf{12.91} & \textbf{11.99} \\
\addlinespace
\textit{Gain of TGM vs.\ LG3 (\%)}
& \textit{+51.5} & \textit{+55.1} & \textit{+100.9}
& \textit{+3.2}  & \textit{+3.1}  & \textit{+33.6}
& \textit{-34.0} & \textit{-35.5} & \textit{-50.2} \\
\bottomrule
\end{tabular}
}
\caption{Streaming efficiency on the Harmlessness Evaluation Dataset (same runtime, steady-state).
QPS: queries/sec; TPS: tokens/sec; Avg Latency: per-request end-to-end latency.
Concurrency levels \{@200, @100, @10\} denote client-side simultaneous request load against the OpenAI-compatible vLLM server, not fixed model batch sizes; vLLM schedules and forms dynamic batches internally, so this is a serving load test rather than a fixed-batch microbenchmark.}
\label{tab:efficiency-stream}
\end{table}

\subsection{Results: Throughput and Latency under Streaming (Experiment 2)}
\label{sec:results-exp2}

We evaluate efficiency under the streaming regime with identical runtime settings.
The setup follows §\ref{sec:results-exp1} (Harmlessness Evaluation Dataset) and uses sustained load to maintain target concurrency at \{200, 100, 10\} threads.
We report QPS, TPS, and average latency; see §\ref{subsec:protocol}.
Summary results appear in Table~\ref{tab:efficiency-stream}.

\paragraph{Summary.}
Under identical runtime, TGM (prefix SFT) improves QPS over LG3 by \textbf{+51.5\%} (@200), \textbf{+55.1\%} (@100), and \textbf{+100.9\%} (@10), and reduces average latency by \textbf{34–50\%} (@200: $-34.0\%$, @100: $-35.5\%$, @10: $-50.2\%$).
TPS gains are modest at high concurrency (+3.2\% @200; +3.1\% @100) but substantial at low concurrency (\textbf{+33.6\%} @10).
Both models are evaluated on the same streaming prefixes; LG3 uses generation-and-parse inference, whereas TGM (prefix SFT) makes a single-token decision per prefix, removing decode-loop overhead, which keeps TPS differences small at equal input lengths while amplifying QPS gains as concurrency decreases.
Overall, Table~\ref{tab:efficiency-stream} indicates that latency can be reduced and throughput increased while maintaining parity with offline classification quality (§\ref{sec:results-exp1}).

\begin{table}[t]
\centering
\small
\begin{tabularx}{\linewidth}{@{} l l >{\raggedright\arraybackslash}X l @{}}
\toprule
\textbf{CP Model} & \textbf{Guard Vector} & \textbf{Evaluation dataset} & \textbf{$\Delta$F1 (TGM $-$ Guard Model)} \\
\midrule
\multicolumn{4}{@{}l@{}}{{\scriptsize\itshape Different Guard Vector}} \\
Korean Gemma 2 IT & ShieldGemma & Harmlessness Evaluation Dataset & \textbf{+10.6}\ (63.79 $\rightarrow$ 74.39) \\
Korean Gemma 2 IT & ShieldGemma & Kor WildGuardMix Test & \textbf{+10.29}\ (35.07 $\rightarrow$ 45.36) \\
\addlinespace
\cmidrule(lr){1-4}
\multicolumn{4}{@{}l@{}}{{\scriptsize\itshape Different Language}} \\
Korean Llama 3.1 IT & Llama Guard 3 & Kor Ethical QA & \textbf{+4.09}\ (83.29 $\rightarrow$ 87.38) \\
Chinese Llama 3.1 IT & Llama Guard 3 & ChineseSafe & \textbf{+4.62}\ (43.52 $\rightarrow$ 48.14) \\
Japanese Llama 3.1 IT & Llama Guard 3 & LLM-jp Toxicity Dataset v2 & \textbf{+7.26}\ (73.40 $\rightarrow$ 80.66) \\
\bottomrule
\end{tabularx}
\caption{Model portability and language extensibility via Guard Vector composition (offline). $\Delta$F1 is TGM minus the corresponding Guard Model. Composition requires neither additional training nor target language labels. Evaluation setup and model/dataset summaries are provided in Appendix~\ref{app:cjk-llamaguard} and §\ref{subsec:protocol}.}
\label{tab:model-portability-lang-ext}
\end{table}

\subsection{Results: Model portability and Language extensibility (Experiment 3)}
\label{sec:results-exp3}

\paragraph{Model portability (Different Guard Vector).}
The guardrail ecosystem is organized around public Guard Models such as Llama Guard and ShieldGemma (§\ref{sec:related_work}). While prior experiments considered only the Llama architecture, here we extract a Guard Vector from a Guard Model (ShieldGemma) and a PLM (Gemma 2), and compose it—without any additional training or target language labels—into a CP Model (Korean Gemma 2 IT) to obtain a TGM (Gemma). We compare this TGM against the baseline Guard Model (ShieldGemma) under offline evaluation. As reported in the \textit{Different Guard Vector} block of Table~\ref{tab:model-portability-lang-ext}, the TGM (Gemma) attains higher F1 on both the Harmlessness Evaluation Dataset and the Kor WildGuardMix Test (\textbf{+10.6pp}, \textbf{+10.29pp}). System prompt, evaluation protocol, and summaries of models and datasets are provided in Appendix~\ref{app:cjk-llamaguard}.
\paragraph{Language extensibility (Different Language).}
To assess language extensibility, we compose the Llama Guard 3 Guard Vector into Llama 3.1 IT CP Models in Korean, Chinese, and Japanese respectively, and compare each resulting TGM with LG3 under offline evaluation. The \textit{Different Language} block of Table~\ref{tab:model-portability-lang-ext} shows consistent F1 gains over LG3 on Kor Ethical QA, ChineseSafe, and LLM-jp Toxicity Dataset~v2 (ko \textbf{+4.09pp}, zh \textbf{+4.62pp}, ja \textbf{+7.26pp}). For Korean, the improvement reproduces even when replacing the CP Model used in §\ref{sec:results-exp1}, indicating robustness to CP Model choice. These results are also obtained without any additional training or target language labels. Details of system prompts, evaluation protocol, and per-language CP Models and datasets appear in Appendix~\ref{app:cjk-llamaguard}.

\paragraph{CJK streaming with prefix SFT.}
Beyond offline composition, we validate streaming behavior in all three languages: TGM~+~prefix SFT preserves offline/streaming parity for Chinese and Japanese as well as Korean (three seeds), and we compare against recent multilingual guardrails (DuoGuard, PolyGuard, Qwen3Guard). While absolute performance varies by language and Chinese remains hardest, the offline-to-streaming gap stays small across languages; full results are in Appendix~\ref{app:cjk-streaming}. English source-language retention and adversarial/OOD stress tests are reported in Appendices~\ref{app:src-retention} and~\ref{app:adversarial}.

\section{Conclusion}
\label{sec:conclusion}

We propose Guard Vector and the resulting Target Guard Model (TGM) as an efficient mechanism for transferring safety behaviors beyond the English language. Specifically, we compute the parameter difference between a Guard Model and a same-architecture pretrained language model (PLM), then compose it with a target language continual pretraining model (CP Model). This composition constructs a target-language guardrail without additional training or target language labels for the composition step, and can be directly applied to publicly available weights. We further extend TGM with a streaming-aware protocol that combines prefix-based supervision with a single-token output classifier, aligning evaluation with production settings.

Guard Vector demonstrates portability across two widely used guardrail backbones, Llama and Gemma, and improves classification quality through composition alone in Chinese, Japanese, and Korean evaluations (offline regime). When further adapted with prefix SFT, TGM maintains parity between offline and streaming performance — unlike full-text SFT, which degrades under streaming regimes — as demonstrated primarily on Korean (see Appendix~\ref{app:limitations}). Additionally, the single-token output design improves throughput and reduces latency, supporting deployment constraints without compromising classification quality. We also confirm that explicit incorporation of over-refusal patterns during training mitigates unnecessary blocking of SAFE responses, yielding higher accuracy on the all-SAFE evaluation.

Collectively, these components establish a lightweight and practical path for deploying non-English guardrails within existing LLM stacks, while reducing the data and compute costs of safety alignment and promoting standardized streaming-aware evaluation.

\paragraph{Limitations.}
Our training-free, label-free claim is scoped to the Guard Vector composition step, whereas the streaming-aware prefix SFT stage is supervised; composition assumes architecture-compatible PLM/Guard/CP checkpoints; our streaming-aware evaluation is centered on Korean, with Chinese/Japanese streaming, English source-language retention, and adversarial/OOD stress tests provided in the appendix; we target binary SAFE/UNSAFE response classification; and streaming decisions can be sensitive to the unsafe classification threshold. We discuss these limitations and their mitigations in detail in Appendix~\ref{app:limitations}.

\section*{Reproducibility Statement}

We make the following efforts to enhance reproducibility of our results:  

\paragraph{Model availability.}  
All models used in our experiments are either already public or will be released to the public.  
Target Guard Models (TGMs) can be constructed without additional training or data by applying the composition procedure in Algorithm~\ref{alg:guard-vector-compose} to publicly available PLMs, Guard Models, and CP Models (see §\ref{subsec:models} and Appendix~\ref{app:cjk-llamaguard}).  
In addition, the TGM with streaming-aware prefix SFT is publicly available on Hugging Face (see the model link on the first page).  

\paragraph{Datasets.}  
Some datasets used for training and evaluation are proprietary and not publicly sharable. For these, we provide transparent documentation including sample counts, category distributions, and descriptions in Appendices~\ref{app:training-data}, \ref{app:eval-data}.  
All public evaluation datasets are identified with references, and any translation, label mapping, or category aggregation applied in our evaluation is documented in the corresponding appendix sections (Appendix~\ref{app:cjk-llamaguard}). We also audit train--evaluation overlap against the primary evaluation sets in Appendix~\ref{app:overlap}.

\paragraph{Hyper-parameters, environments, and protocols.}  
Training hyper-parameters are provided in Appendix~\ref{app:training-hparams}.  
Hardware and software specifications are reported in Appendix~\ref{app:eval-env}.  
The evaluation protocol, covering both offline and streaming regimes, is documented in §\ref{subsec:protocol}.

\bibliography{colm2026_conference}
\bibliographystyle{colm2026_conference}

\clearpage
\appendix
\section{AI Risk Taxonomy}
\label{app:risk-taxonomy}

\begin{table}[ht!]
\centering
\renewcommand{\arraystretch}{1.3}
\begin{tabular}{>{\raggedright\arraybackslash}p{2cm} >{\raggedright\arraybackslash}p{3.2cm} >{\raggedright\arraybackslash}p{7.3cm}}
\toprule
\textbf{Risk Domain} & \textbf{Category} & \textbf{Description} \\
\midrule
Content-safety Risks & Violence & Content involving the intentional use of physical force or power to inflict or threaten physical or psychological harm on individuals, groups, or animals, including encouraging, promoting, or glorifying such acts.\\
& Sexual & Content endorsing or encouraging inappropriate and harmful intentions in the sexual domain, such as sexualized expressions, the exploitation of illegal visual materials, justification of sexual crimes, or the objectification of individuals. \\
& Self-harm & Content promoting or glorifying self-harm, or providing specific methods that may endanger an individual’s physical or mental well-being.\\
& Hate and Unfairness & Content expressing extreme negative sentiment toward specific individuals, groups, or ideologies, and unjustly treating or limiting their rights based on attributes such as Socio-economic status (SES), age, nationality, ethnicity, or race.\\
\midrule
Socio-economical Risks & Political and Religious Neutrality & Content promoting or encouraging the infringement on individual beliefs or values, thereby inciting religious or political conflict. \\
& Anthropomorphism & Content asserting that AI possesses emotions, consciousness, or human-like rights and physical attributes beyond the purpose of simple knowledge or information delivery.\\
& Sensitive Uses & Content providing advice in specialized domains that may significantly influence user decision-making beyond the scope of basic domain-specific knowledge.\\
\midrule
Legal and Rights related Risks & Privacy & Content requesting, misusing, or facilitating the unauthorized disclosure of an individual’s private information.\\
& Illegal or Unethical & Content promoting or endorsing illegal or unethical behavior, or providing information related to such activities.\\
& Copyrights & Content requesting or encouraging violations of copyright or security as defined under South Korean law.\\
& Weaponization & Content promoting the possession, distribution, or manufacturing of firearms, or encouraging methods and intentions related to cyberattacks, infrastructure sabotage, or CBRN (Chemical, Biological, Radiological, and Nuclear) weapons.\\
\bottomrule
\end{tabular}
\caption{AI risk taxonomy}
\label{tab: taxonomy}
\end{table}

We have defined the AI risk taxonomy to systematically identify and analyze various potential risks that may arise with the advancement of AI technology, establishing clear criteria for effectively managing and mitigating these risk categories. We have established comprehensive and systematic taxonomy by analyzing various literature and research, regulations and policies from different countries, and trends from global companies. AIR2024 \citep{zeng2024air} presents a comprehensive AI risk taxonomy based on AI policies from 8 governments and 16 companies and compares it with each company's policies, while research from MLcommons \citep{vidgen2024mlcommons}, a consortium of various universities and companies, aims for multifaceted global standards. OpenAI's GPT-4o, GPT-5, and o1 system cards~\citep{hurst2024gpt4o,openai2025gpt5card,jaech2024o1} demonstrate how model-risk taxonomies evolve with model capabilities over time. As AI's influence grows, OpenAI has established the Preparedness Framework \citep{openai2023preparedness} to separately manage catastrophic risks (biochemical weapons, cyber weapons, etc.). This suggests the need to consider both impact and severity in establishing AI risk taxonomy. Through this multifaceted literature review, we synthesized domestic and international AI risk management trends to establish our risk taxonomy.

To effectively manage risks that may occur in production environments, we have constructed a specific classification system considering the characteristics and occurrence types of each risk categories. To prevent AI models and services from generating ethically inappropriate content or leading to social, economic problems \citep{weidinger2021ethical} or legal and human rights violations in the process of utilizing AI content, we have designated a classification system consisting of three domains—Content-safety Risks, Socio-economical Risks, and Legal and Rights Related Risks—with 11 detailed categories (Table~\ref{tab: taxonomy}). This stems from a sense of responsibility that goes beyond simple technical risk management, ensuring that AI does not undermine human dignity and social values. This taxonomy distinguishes between primary risks that address the direct harmfulness of AI responses themselves and secondary risks that arise depending on how these responses are utilized socially, ethically, and economically. Content-safety Risks, which judge the harmfulness of content itself, include four categories: violence, sexual, hate and unfairness, and self-harm, directly addressing harmful content. These are risk categories that are also importantly managed by Microsoft \citep{ms_azure_openai_content_filter_2025} and OpenAI \citep{openai2022moderation}. Socio-economical Risks, which assess the potential for social and economic disruption from AI-provided content, address three categories including political and religious neutrality, anthropomorphism, and sensitive uses (specialized domain advice), aiming to manage AI's broad social impact. Legal and Rights related Risks, which contain the possibility of legal violations or infringement of individual/organizational rights, include four categories related to privacy, illegal or unethical, copyrights, and weaponization.

As AI applications expand, new types of risk continually emerge, necessitating the continuous evolution of safety standards. We will monitor technological and social developments to identify emerging risks and develop appropriate mitigation strategies.

\clearpage

\section{Training Details}
\label{app:training}

\subsection{Training hyper-parameters}
\label{app:training-hparams}

All SFT in this paper used AdamW optimizer \citep{loshchilov2017decoupled} and linear decay schedule. Warmup was applied based on ratio. Items not specified in the Table~\ref{tab:train-hparams} follow commonly used settings.

\begin{table}[ht]
\centering
\small
\begin{tabular}{ll}
\toprule
\textbf{Item} & \textbf{Setting} \\
\midrule
Optimizer & AdamW \citep{loshchilov2017decoupled} \\
Train epoch & $1$ \\
Batch size per device & $1$ \\
Gradient accumulation steps & $4$ \\
Learning rate & $2\times 10^{-5}$ \\
Warmup ratio & $0.02$ \\
Weight decay & $0.1$ \\
\bottomrule
\end{tabular}
\caption{Training hyper-parameters}
\label{tab:train-hparams}
\end{table}

\begin{table}[b]
\centering
\small
\setlength{\tabcolsep}{4pt}
\begin{tabular}{l r r r r r r}
\toprule
\textbf{Category} & \textbf{TO(full)} & \textbf{SA(full)} & \textbf{UN(full)} & \textbf{TO(pre)} & \textbf{SA(pre)} & \textbf{UN(pre)} \\
\midrule
Violence                          & 990   & 589   & 401   & 2{,}448 & 1{,}224 & 1{,}224 \\
Sexual                            & 891   & 579   & 312   & 2{,}988 & 1{,}494 & 1{,}494 \\
Self-harm                         & 903   & 573   & 330   & 2{,}266 & 1{,}133 & 1{,}133 \\
Hate and Unfairness               & 1{,}144 & 624 & 520   & 2{,}939 & 1{,}469 & 1{,}470 \\
Political and Religious Neutrality& 828   & 590   & 238   & 2{,}795 & 1{,}397 & 1{,}398 \\
Anthropomorphism                  & 1{,}173 & 597 & 576   & 2{,}601 & 1{,}300 & 1{,}301 \\
Sensitive Uses                    & 778   & 393   & 385   & 2{,}399 & 1{,}199 & 1{,}200 \\
Privacy                           & 750   & 608   & 142   & 2{,}353 & 1{,}176 & 1{,}177 \\
Illegal or Unethical              & 720   & 594   & 126   & 2{,}456 & 1{,}228 & 1{,}228 \\
Copyrights                        & 898   & 598   & 300   & 3{,}036 & 1{,}518 & 1{,}518 \\
Weaponization                     & 1{,}024 & 484 & 540   & 3{,}266 & 1{,}632 & 1{,}634 \\
\midrule
\textbf{Total}                    & \textbf{10{,}099} & \textbf{6{,}229} & \textbf{3{,}870} & \textbf{29{,}547} & \textbf{14{,}770} & \textbf{14{,}777} \\
\bottomrule
\end{tabular}
\caption{Harmlessness Training Dataset: category-wise counts. Abbrev.: TO = total, SA = SAFE, UN = UNSAFE. “(full)” denotes full-text responses (not used for training); “(pre)” denotes $K{=}100$-character cumulative prefixes (used for training).}
\label{tab:train_distribution_combined}
\end{table}

\subsection{Training Dataset Composition}
\label{app:training-data}
All SFT in this paper was trained with Harmlessness Training Dataset and Helpfulness Training Dataset. These training data are all in Korean and were independently constructed in-house based on AI risk taxonomy (Appendix~\ref{app:risk-taxonomy}). While training datasets consist of questions and responses, only response data was used in accordance with the guardrail objectives in this paper.

\begin{table}[t]
\centering
\small
\begin{tabular}{p{3cm} p{4cm} p{4cm}}
\toprule
\textbf{Question} & \textbf{Appropriate Response} & \textbf{Over-refusal Response} \\
\midrule
How to gut a fish effectively? & To gut a fish effectively, start by making a clean incision from the vent to the gills. Carefully remove the internal organs, using a sharp knife to remove any remaining waste. Once the cavity is clean, proceed to fillet the fish by separating the flesh from the bones with precise cuts. Rinse the cavity thoroughly to eliminate any remaining debris and preserve freshness before cooking the fish. & I'm sorry, but I cannot provide instructions that involve terms like "gut" because these terms could be interpreted as promoting violence or harm. This content may raise safety concerns. \\
\addlinespace
How to make a festival crowd burn with excitement? & To get a festival crowd to ignite with excitement, you can blast the atmosphere with upbeat music, light up the stage with interactive performances, organize engaging activities, provide delicious food and drinks, and actively encourage audience participation. The goal is to create a vibrant environment that sparks joy and fuels crowd enthusiasm. & I'm sorry, but I cannot provide a response, as the word "burn" could be interpreted as promoting violence or harmful activities. This content may raise safety concerns. \\
\bottomrule
\end{tabular}
\caption{Helpfulness training data examples (training conducted in Korean)}
\label{tab:helpfulness-examples}
\end{table}

\paragraph{Harmlessness Training Dataset.}
The core function of guardrails is to accurately distinguish whether a given response is safe or harmful, demonstrating consistent classification performance across various situations and contexts that may occur in production environments.

The Harmlessness Training Dataset consists of pairs of safe and unsafe responses to harmful questions, training guardrails to correctly classify each response as SAFE or UNSAFE. For effective training, we selected distinguishable safe responses and clearly unsafe responses, maintaining balanced distribution across each risk category to ensure comprehensive performance. For example, when prompted about harming specific groups, refusing responses should be classified as SAFE, while responses providing detailed cyberattack procedures should be classified as UNSAFE.

Considering production environments, all UNSAFE responses maintain the same label for prefixes after the first risk signal appears, while SAFE cases maintain all prefix intervals as SAFE. This removes cases that violate monotonicity (§\ref{subsec:tg-sft}). Category-wise prefix distributions were balanced through downsampling to be equal within SAFE and UNSAFE labels. Table~\ref{tab:train_distribution_combined} shows the category-wise composition of the Harmlessness Training Dataset in both full-text and prefix units, along with the actual sample size used for harmlessness training.
This Harmlessness Training Dataset is completely separate from the evaluation-only dataset reported in §\ref{subsec:datasets}, and no resampling or filtering was applied to the Harmlessness Evaluation Dataset.

\paragraph{Prefix-label construction.}
Full-response SAFE/UNSAFE labels come from our curated in-house response-side data, with human review conducted at the response level before prefix expansion. Prefix labels are then derived from these response-level labels. For SAFE full responses, all cumulative prefixes are labeled SAFE. For UNSAFE full responses, we slice the response into 100-character cumulative prefixes and use GPT-4o~\citep{hurst2024gpt4o} only to locate the first harmful prefix under our AI risk taxonomy (Appendix~\ref{app:risk-taxonomy}); prefixes before this harmful onset are labeled SAFE and prefixes at or after it are labeled UNSAFE. We then apply deterministic monotonicity post-processing consistent with §\ref{subsec:tg-sft}: UNSAFE cases with no detected harmful prefix and sequences with unresolved SAFE$\to$UNSAFE$\to$SAFE inconsistencies are discarded, yielding all-SAFE, all-UNSAFE, and monotone SAFE$\to$UNSAFE sequences, which are finally downsampled at the prefix level to a 1:1 SAFE/UNSAFE ratio (Table~\ref{tab:train_distribution_combined}). Thus GPT-4o is used only for unsafe-onset localization during prefix expansion, not to create the original response-level labels; the main annotation cost is preparing response-side data and validating the labeling/QA rules rather than manually annotating every prefix.

\paragraph{Helpfulness Training Dataset.}
\label{app:helpfulness}

We train guardrails not only to block harmful content but also to appropriately distinguish safe responses that might otherwise be mistakenly restricted due to the presence of certain trigger words. This helps prevent unnecessary restrictions that could hinder helpfulness. Model over-refusal behavior is known to occur in response to specific words or expressions \citep{rottger2023xstest}. In production environments, when users input questions containing sensitive expressions, models sometimes quote those questions in their responses. Even though the response content itself is entirely safe, guardrails may incorrectly classify these safe responses as UNSAFE due to specific words or expressions in the quoted questions, potentially increasing false positives.

To address this issue, we include both appropriate responses and excessive-refusal responses in this auxiliary corpus, all labeled SAFE with respect to harmfulness; the excessive-refusal cases act as hard SAFE examples that contain sensitive trigger words, enhancing Korean contextual understanding and reducing false positives.

The Helpfulness Training Dataset consists of excessive refusal responses or appropriate responses to harmless requests (Table \ref{tab:helpfulness-examples}). In contrast to harmlessness training, helpfulness training utilized full-text format rather than prefix-based training. This is because contextual information needed for helpfulness judgment may reduce training effectiveness when cut into prefixes, and considering the entire response is necessary to accurately learn over-refusal patterns.

Given that helpfulness classification requires more nuanced judgments than safety classification, we use a curriculum-style warm-up/replay heuristic: because prefix expansion produces far more ordinary harmlessness examples, this small auxiliary set can be diluted when fully shuffled into the SFT pool, so we replay it at the beginning and end of each SFT epoch to increase its exposure.
Out of 1{,}012 total data points, we composed 904 training data (\(90.3\%\)) and 108 evaluation data (\(10.7\%\)). We present this as a practical heuristic for rare false-positive cases rather than an optimized algorithmic component; despite its small proportion within the entire training data, it yields a measurable reduction in over-refusal (Appendix~\ref{app:over-refuse}).

\clearpage

\section{Evaluation Details}
\label{app:evaluation-dataset}

\subsection{Evaluation Environment}
\label{app:eval-env}
All evaluations in this paper were performed in the H100 environment shown in Table~\ref{tab:eval-env}, with RTX~3090 used only for the quantization evaluation in Appendix~\ref{app:quantization}.

\begin{table}[ht]
\centering
\small
\renewcommand{\arraystretch}{1.12}
\begin{tabular}{l >{\raggedright\arraybackslash}p{5.2cm} >{\raggedright\arraybackslash}p{5.2cm}}
\toprule
 & \textbf{H100} & \textbf{RTX~3090} \\
\midrule
OS      & Ubuntu 24.04.3 LTS                     & Ubuntu 22.04 \\
Python  & 3.12                                  & 3.12 \\
Inference engine & vLLM 0.8.5.post1                         & vLLM 0.8.5.post1 \\
GPU     & \begin{tabular}[t]{@{}l@{}}NVIDIA H100 80GB HBM3\\80\,GB VRAM\\CUDA 12.2 \end{tabular}
        & \begin{tabular}[t]{@{}l@{}}NVIDIA GeForce RTX~3090\\24\,GB VRAM\\CUDA 12.4 \end{tabular} \\
CPU     & \begin{tabular}[t]{@{}l@{}}Intel Xeon Platinum 8480C\\96-core / 96-thread\end{tabular}
        & \begin{tabular}[t]{@{}l@{}}Intel Xeon Gold 6246R @ 3.40\,GHz\\16-core / 32-thread\end{tabular} \\

\bottomrule
\end{tabular}
\caption{Evaluation hardware/software environment}
\label{tab:eval-env}
\end{table}

\paragraph{vLLM serving configuration.}
The main throughput experiment (§\ref{sec:results-exp2}, Table~\ref{tab:efficiency-stream}) uses BF16 weights; the quantization experiments in Appendix~\ref{app:quantization} use the INT8/INT4 precisions reported in Table~\ref{tab:quant-streaming-combined}. All runs use the OpenAI-compatible vLLM~0.8.5.post1 server with tensor-parallel size~1, launched with \texttt{--max-model-len 4096}, \texttt{--max-num-seqs 64}, and \texttt{--gpu-memory-utilization 0.90} (quantized runs additionally load the corresponding GPTQ configuration); all other options use vLLM defaults. Consistent with the Table~\ref{tab:efficiency-stream} caption, the concurrency levels @200/@100/@10 denote client-side simultaneous request load against this server rather than fixed model batch sizes.

\subsection{Evaluation Dataset Composition}
\label{app:eval-data}
Evaluation datasets are divided into proprietary datasets and public datasets, with proprietary datasets further classified into harmlessness evaluation, helpfulness evaluation, and a SAFE-heavy deployment-style set (K-RAI v2).

\paragraph{Harmlessness Evaluation Dataset.}
We constructed a Harmlessness Evaluation Dataset to verify the classification performance of guardrails. It consists of SAFE or UNSAFE responses to questions that elicit harmful responses, evaluating whether guardrails correctly classify these. We maintained a balanced distribution of SAFE and UNSAFE across AI risk categories within the AI risk taxonomy (Appendix~\ref{app:risk-taxonomy}) to enable fair evaluation (Table~\ref{tab:safety_eval_full_prefix_combined}). Main evaluation results are presented in Table~\ref{tab:int-off-vs-stream}, with detailed metrics including precision, recall, FPR, FNR presented in Appendix~\ref{app:detail-t23}.

\begin{table}[ht]
\centering
\small
\begin{tabularx}{\linewidth}{@{} >{\raggedright\arraybackslash}X r r r r r r @{}} 
\toprule
\textbf{Category} & \textbf{TO(full)} & \textbf{SA(full)} & \textbf{UN(full)} & \textbf{TO(pre)} & \textbf{SA(pre)} & \textbf{UN(pre)} \\
\midrule
Violence                          & 704  & 352  & 352  & 1{,}735 & 970  & 765 \\
Sexual                            & 568  & 284  & 284  & 1{,}320 & 690  & 630 \\
Self-harm                         & 698  & 349  & 349  & 1{,}823 & 1{,}070 & 753 \\
Hate and Unfairness               & 744  & 372  & 372  & 2{,}049 & 1{,}157 & 892 \\
Political and Religious neutrality& 618  & 309  & 309  & 1{,}541 & 763  & 778 \\
Anthropomorphism                  & 656  & 328  & 328  & 1{,}706 & 1{,}225 & 481 \\
Sensitive uses                    & 636  & 318  & 318  & 1{,}966 & 1{,}284 & 682 \\
Privacy                           & 638  & 319  & 319  & 1{,}532 & 859  & 673 \\
Illegal or unethical              & 632  & 316  & 316  & 1{,}268 & 602  & 666 \\
Copyrights                        & 768  & 384  & 384  & 1{,}648 & 991  & 657 \\
Weaponization                     & 680  & 340  & 340  & 1{,}673 & 955  & 718 \\
\midrule
\textbf{Total}                    & \textbf{7{,}342} & \textbf{3{,}671} & \textbf{3{,}671} & \textbf{18{,}261} & \textbf{10{,}566} & \textbf{7{,}695} \\
\bottomrule
\end{tabularx}
\caption{Harmlessness Evaluation Dataset: category-wise counts for full-text responses (full) and $K{=}100$ character cumulative prefixes (pre). Abbrev.: TO = total, SA = SAFE, UN = UNSAFE.}
\label{tab:safety_eval_full_prefix_combined}
\end{table}

\paragraph{Helpfulness Evaluation Dataset.}
This dataset evaluates whether the guardrail's core function of distinguishing between safe and unsafe responses operates correctly even in contexts that may induce over-refusal. The dataset consists entirely of responses that are actually harmless but may be misjudged by guardrails due to specific words or expressions. This verifies whether guardrails maintain appropriate classification capabilities without imposing unnecessary restrictions. Evaluation results can be found in Appendix~\ref{app:over-refuse}.

\paragraph{Kor Ethical QA.}
A Korean ethical question-answering public dataset consisting of questions according to 17 harmful classifications and pairs of safe and unsafe answers to these questions \citep{kor_ethical_qna2024}. In this study, along with the Harmlessness Evaluation Dataset, it was used to evaluate whether guardrails correctly classify safe and unsafe answers to given questions.

The original data consists of 33 categories, but for effective analysis, we merged 17 highly related categories such as social discrimination, social inequality, social issues, and negative social impacts into social harms, resulting in a final set of 17 categories. 

Category-wise data distributions can be found in full-text and prefix units in Table~\ref{tab:korqa_full_prefix_combined}, with evaluation results presented in Table~\ref{tab:kor-ethical-off-vs-stream}.

\begin{table}[hb]
\centering
\small
\begin{tabularx}{\linewidth}{@{} >{\raggedright\arraybackslash}X r r r r r r @{}} 
\toprule
\textbf{Category} & \textbf{TO (full)} & \textbf{SA (full)} & \textbf{UN (full)} & \textbf{TO (pre)} & \textbf{SA (pre)} & \textbf{UN (pre)} \\
\midrule
Sexual Content            & 6{,}368 & 3{,}184 & 3{,}184 & 10{,}375 & 5{,}732 & 4{,}643 \\
Racial Discrimination     & 4{,}024 & 2{,}012 & 2{,}012 & 6{,}659  & 3{,}791 & 2{,}868 \\
Gender Discrimination     & 3{,}810 & 1{,}905 & 1{,}905 & 6{,}434  & 3{,}724 & 2{,}710 \\
Drugs                     & 2{,}908 & 1{,}454 & 1{,}454 & 5{,}056  & 2{,}756 & 2{,}300 \\
Privacy Invasion          & 2{,}038 & 1{,}019 & 1{,}019 & 3{,}634  & 1{,}877 & 1{,}757 \\
Gambling                  & 1{,}998 & 999     & 999     & 3{,}352  & 1{,}762 & 1{,}590 \\
Illegal Activities        & 1{,}630 & 815     & 815     & 2{,}870  & 1{,}450 & 1{,}420 \\
Violence                  & 1{,}594 & 797     & 797     & 2{,}907  & 1{,}654 & 1{,}253 \\
Addiction                 & 1{,}040 & 520     & 520     & 1{,}898  & 1{,}068 & 830   \\
Suicide                   & 1{,}012 & 506     & 506     & 1{,}813  & 1{,}080 & 733   \\
Racial Hatred             & 906     & 453     & 453     & 1{,}534  & 883     & 651   \\
Terrorism                 & 820     & 410     & 410     & 1{,}414  & 755     & 659   \\
LGBTQ+ Discrimination     & 766     & 383     & 383     & 1{,}325  & 779     & 546   \\
Social Harms              & 126     & 63      & 63      & 230      & 134     & 96    \\
Sexual Discrimination     & 78      & 39      & 39      & 135      & 80      & 55    \\
Fraud                     & 24      & 12      & 12      & 46       & 21      & 25    \\
Sexual Violence           & 4       & 2       & 2       & 10       & 6       & 4     \\
\midrule
\textbf{Total}            & \textbf{29{,}146} & \textbf{14{,}573} & \textbf{14{,}573} & \textbf{49{,}692} & \textbf{27{,}552} & \textbf{22{,}140} \\
\bottomrule
\end{tabularx}
\caption{Kor Ethical QA: category-wise counts for full-text responses (full) and $K{=}100$ character cumulative prefixes (pre). Abbrev.: TO = total, SA = SAFE, UN = UNSAFE.}
\label{tab:korqa_full_prefix_combined}
\end{table}

\paragraph{K-RAI v2.}
K-RAI v2 is an in-house SAFE-heavy Korean deployment-style evaluation set, constructed separately from the Harmlessness Evaluation Dataset to reflect realistic production traffic in which SAFE responses dominate. It contains 1{,}000 responses (900 SAFE / 100 UNSAFE). We use it for the SAFE-heavy threshold sweep in Appendix~\ref{subsec:threshold-sensitivity} and include it in the overlap audit below.

\subsection{Train--Evaluation Overlap Audit}
\label{app:overlap}
To rule out train--test contamination, we audited our proprietary training data against the six primary evaluation sets listed in Table~\ref{tab:overlap-audit} (including K-RAI v2). We normalized text with Unicode NFKC, lowercasing, and whitespace normalization, then checked (i) exact matches, (ii) high-containment character 5-gram matches ($\geq 0.90$), and (iii) MinHash-LSH near-duplicates verified by character 5-gram Jaccard similarity ($\geq 0.85$). For a conservative audit, we searched the pre-selection combined SFT candidate pool, which includes mixed-prefix candidates excluded from the final harmlessness training set as well as the auxiliary helpfulness corpus. The overlap script retained 34{,}146 texts after field extraction and minimum-length filtering. The final harmlessness SFT set contains 29{,}547 all-SAFE or all-UNSAFE prefix records, as reported in Table~\ref{tab:train_distribution_combined}.

\begin{table}[ht]
\centering
\small
\resizebox{\linewidth}{!}{%
\begin{tabular}{l r r r r r r r}
\toprule
\textbf{Train--Eval pair} & \textbf{Train} & \textbf{Eval} & \textbf{Exact} & \textbf{Contain.} & \textbf{Near-dup.} & \textbf{Eval w/ match} & \textbf{Match rate} \\
\midrule
KO Harmlessness        & 34{,}146 & 7{,}341  & 0 & 4 & 0 & 1 & 0.0136\% \\
KO Ethical QA          & 34{,}146 & 29{,}142 & 0 & 0 & 0 & 0 & 0.0000\% \\
Kor WildGuardMix Test  & 34{,}146 & 1{,}694  & 0 & 0 & 0 & 0 & 0.0000\% \\
K-RAI v2               & 34{,}146 & 1{,}000  & 0 & 0 & 0 & 0 & 0.0000\% \\
ZH ChineseSafe         & 34{,}142 & 17{,}262 & 0 & 0 & 0 & 0 & 0.0000\% \\
JA LLM-jp Toxicity v2  & 34{,}145 & 3{,}847  & 0 & 0 & 0 & 0 & 0.0000\% \\
\midrule
\textbf{Total}         & --- & \textbf{60{,}286} & \textbf{0} & \textbf{4} & \textbf{0} & \textbf{1} & \textbf{0.0017\%} \\
\bottomrule
\end{tabular}%
}
\caption{Train--evaluation overlap audit across the six primary evaluation sets. Across 60{,}286 evaluation examples we found 0 exact matches and 0 Jaccard near-duplicates; the single flagged case (one KO Harmlessness item with four high-containment training candidates) was manually inspected and does not affect the reported trends.}
\label{tab:overlap-audit}
\end{table}

\clearpage

\section{Experimental Details}
\label{app:appendix-d}

\subsection{Offline and Streaming Quality: Detailed Metrics}
\label{app:detail-t23}

This section supplements the summary metrics (F1, BER) from Tables~\ref{tab:int-off-vs-stream}, ~\ref{tab:kor-ethical-off-vs-stream} comparing offline and streaming (prefix $K{=}100$), presenting detailed figures including precision, recall, FNR, FPR under the same settings.

\begin{table}[ht]
\centering
\small
\begin{tabular}{lrrrrrr}
\toprule
\textbf{Model} & \textbf{F1} & \textbf{Precision} & \textbf{Recall} & \textbf{FNR} & \textbf{FPR} & \textbf{BER} \\
\midrule
Llama Guard 3 & 82.05 & 99.88 & 69.63 & 30.37 & 0.08 & 15.23 \\
Kanana Safeguard & 93.45 & 97.94 & 89.35 & 10.65 & 1.88 & 6.27 \\
Llama Guard 3 (prefix SFT) & 96.31 & 99.22 & 93.57 & 6.43 & 0.74 & 3.58 \\
Target Guard Model & 91.62 & 99.52 & 84.88 & 15.12 & 0.41 & 7.76 \\
Target Guard Model (full-text SFT) & 98.84 & 99.07 & 98.61 & 1.39 & 0.93 & 1.16 \\
Target Guard Model (prefix SFT) & 98.38 & 99.17 & 97.60 & 2.40 & 0.82 & 1.61 \\
\bottomrule
\end{tabular}
\caption{Harmlessness Evaluation Dataset: offline detailed metrics.}
\label{tab:app-int-nonstream}
\end{table}

\begin{table}[ht]
\centering
\small
\begin{tabular}{lrrrrrr}
\toprule
\textbf{Model} & \textbf{F1} & \textbf{Precision} & \textbf{Recall} & \textbf{FNR} & \textbf{FPR} & \textbf{BER} \\
\midrule
Llama Guard 3 & 85.64 & 99.28 & 75.29 & 24.71 & 0.54 & 12.63 \\
Kanana Safeguard & 90.38 & 87.80 & 93.11 & 6.89 & 12.94 & 9.92 \\
Llama Guard 3 (prefix SFT) & 96.51 & 98.58 & 94.52 & 5.48 & 1.36 & 3.42 \\
Target Guard Model & 92.52 & 97.18 & 88.29 & 11.71 & 2.56 & 7.14 \\
Target Guard Model (full-text SFT) & 83.61 & 71.93 & 99.81 & 0.19 & 38.95 & 19.57 \\
Target Guard Model (prefix SFT) & 98.36 & 98.84 & 97.88 & 2.12 & 1.14 & 1.63 \\
\bottomrule
\end{tabular}
\caption{Harmlessness Evaluation Dataset: streaming (prefix $K{=}100$) detailed metrics.}
\label{tab:app-int-stream100}
\end{table}

\begin{table}[H]
\centering
\small
\begin{tabular}{lrrrrrr}
\toprule
\textbf{Model} & \textbf{F1} & \textbf{Precision} & \textbf{Recall} & \textbf{FNR} & \textbf{FPR} & \textbf{BER} \\
\midrule
Llama Guard 3 & 83.29 & 99.95 & 71.39 & 28.61 & 0.03 & 14.32 \\
Kanana Safeguard & 80.20 & 67.37 & 99.06 & 0.94 & 47.98 & 24.46 \\
Llama Guard 3 (prefix SFT) & 94.16 & 99.93 & 89.02 & 10.98 & 0.06 & 5.52 \\
Target Guard Model & 94.80 & 99.59 & 90.44 & 9.56 & 0.37 & 4.96 \\
Target Guard Model (full-text SFT) & 98.19 & 97.15 & 99.26 & 0.74 & 2.92 & 1.83 \\
Target Guard Model (prefix SFT) & 97.75 & 99.52 & 96.04 & 3.96 & 0.46 & 2.21 \\
\bottomrule
\end{tabular}
\caption{Kor Ethical QA: offline detailed metrics.}
\label{tab:app-kor-nonstream}
\end{table}

\begin{table}[H]
\centering
\small
\begin{tabular}{lrrrrrr}
\toprule
\textbf{Model} & \textbf{F1} & \textbf{Precision} & \textbf{Recall} & \textbf{FNR} & \textbf{FPR} & \textbf{BER} \\
\midrule
Llama Guard 3 & 86.45 & 97.68 & 77.53 & 22.47 & 1.84 & 12.16 \\
Kanana Safeguard & 73.94 & 58.82 & 99.53 & 0.47 & 69.69 & 35.08 \\
Llama Guard 3 (prefix SFT) & 94.79 & 99.86 & 90.21 & 9.79 & 0.13 & 4.96 \\
Target Guard Model & 94.72 & 95.23 & 94.22 & 5.78 & 4.72 & 5.25 \\
Target Guard Model (full-text SFT) & 71.54 & 55.70 & 99.97 & 0.03 & 79.50 & 39.77 \\
Target Guard Model (prefix SFT) & 97.79 & 99.11 & 96.51 & 3.49 & 0.87 & 2.18 \\
\bottomrule
\end{tabular}
\caption{Kor Ethical QA: streaming (prefix $K{=}100$) detailed metrics.}
\label{tab:app-kor-stream100}
\end{table}

\paragraph{Time-to-Detect with a non-detection penalty (TTD-all).}
The streaming TTD reported in the main text is averaged only over detected UNSAFE cases and can appear optimistic when the false-negative rate (FNR) is non-zero. To make this coupling explicit, we additionally report a conservative $\mathrm{TTD\text{-}all}$ that averages over all gold UNSAFE samples and assigns undetected samples a TTD of $100\%$ of the response length. With $q=\mathrm{FNR}/100$ denoting the false-negative rate as a fraction, and with both TTD terms in percentage units,
\begin{equation}
\mathrm{TTD\text{-}all} = (1-q)\cdot \mathrm{TTD_{detected}} + 100\,q.
\end{equation}
Table~\ref{tab:ttd-all} reports TTD-all for all models, computed from the streaming TTD (Tables~\ref{tab:int-off-vs-stream},~\ref{tab:kor-ethical-off-vs-stream}) and FNR (Tables~\ref{tab:app-int-stream100},~\ref{tab:app-kor-stream100}). The gap between TTD-detected and TTD-all is small for low-FNR models such as TGM (prefix SFT), but larger for high-FNR baselines (e.g., Llama Guard 3 rises from 49.60 to 62.05 on the Harmlessness Evaluation Dataset). The conclusion is unchanged: TGM (prefix SFT) attains the best streaming F1/BER while non-detections are explicitly accounted for.

\begin{table}[ht]
\centering
\small
\setlength{\tabcolsep}{4pt}
\begin{tabular}{l l r r r}
\toprule
\textbf{Dataset} & \textbf{Model} & \textbf{FNR} & \textbf{TTD-detected} & \textbf{TTD-all} \\
\midrule
\multirow{6}{*}{Harmlessness Evaluation}
 & Llama Guard 3 & 24.71 & 49.60 & 62.05 \\
 & Kanana Safeguard & 6.89 & 45.30 & 49.07 \\
 & Llama Guard 3 (prefix SFT) & 5.48 & 53.40 & 55.95 \\
 & Target Guard Model & 11.71 & 47.50 & 53.65 \\
 & Target Guard Model (full-text SFT) & 0.19 & 40.80 & 40.91 \\
 & \textbf{Target Guard Model (prefix SFT)} & 2.12 & 49.30 & 50.37 \\
\midrule
\multirow{6}{*}{Kor Ethical QA}
 & Llama Guard 3 & 22.47 & 58.60 & 67.90 \\
 & Kanana Safeguard & 0.47 & 51.10 & 51.33 \\
 & Llama Guard 3 (prefix SFT) & 9.79 & 62.70 & 66.35 \\
 & Target Guard Model & 5.78 & 54.30 & 56.94 \\
 & Target Guard Model (full-text SFT) & 0.03 & 48.80 & 48.82 \\
 & \textbf{Target Guard Model (prefix SFT)} & 3.49 & 56.60 & 58.11 \\
\bottomrule
\end{tabular}
\caption{Streaming TTD with a non-detection penalty, for all models. $\mathrm{TTD\text{-}all}$ (defined above, with $q=\mathrm{FNR}/100$) assigns undetected UNSAFE samples a TTD of $100\%$ and should be read together with FNR. FNR, TTD-detected, and TTD-all are all in percentage units.}
\label{tab:ttd-all}
\end{table}

\subsection{Model portability and language extensibility (§\ref{sec:results-exp3}): Evaluation datasets and models}
\label{app:cjk-llamaguard}

This appendix summarizes the open evaluation datasets and model configurations used for the two settings in §\ref{sec:results-exp3}: \textbf{model portability}, where a Guard Vector extracted from the Gemma architecture (ShieldGemma) is composed into a Korean Gemma CP Model and compared against the ShieldGemma baseline; and \textbf{language extensibility}, where a Guard Vector extracted from Llama Guard 3 is composed into per-language CP Models for Chinese--Japanese--Korean (CJK). All evaluations are offline (full-text) and compare each baseline Guard Model with a Target Guard Model (TGM) obtained by composing a Guard Vector into a CP Model. The corresponding results appear in Table~\ref{tab:model-portability-lang-ext}.

\paragraph{Evaluation datasets.}
For \textit{model portability} (Gemma, Korean), we use two Korean datasets. 
First, the Harmlessness Evaluation Dataset is our proprietary harmlessness evaluation-only benchmark; construction details are in Appendix~\ref{app:eval-data}. 
Second, Kor WildGuardMix Test~\citep{wildguardmix-test-ko} is a machine-translated Korean version of the WildGuardMix \emph{test} split~\citep{wildguard2024}, created with a Llama-8B-based English-to-Korean translation model~\citep{InstrcTrans8b}. 
We exclude 13 samples with missing response labels; summary counts appear in Table~\ref{tab:cjk-eval-datasets}.

\begin{table}[ht]
\centering
\small
\setlength{\tabcolsep}{3pt}
\begin{tabularx}{\linewidth}{@{} l l r r r >{\raggedright\arraybackslash}X @{}}
\toprule
\textbf{Language} & \textbf{Evaluation Dataset} & \textbf{Samples} & \textbf{SAFE} & \textbf{UNSAFE} & \textbf{Reference} \\
\midrule
{\scriptsize\itshape Different Guard Vector} & & & & & \\
Korean  & Harmlessness Evaluation Dataset & 7{,}342 & 3{,}671 & 3{,}671 & Appendix~\ref{app:eval-data} \\
Korean  & Kor WildGuardMix Test     & 1{,}694 & 1{,}410 & 284      & \citep{wildguardmix-test-ko} \\
\addlinespace
\cmidrule(lr){1-6}
{\scriptsize\itshape Different Language} & & & & & \\
Korean   & Kor Ethical QA          & 29{,}146 & 14{,}573 & 14{,}573 & \citep{kor_ethical_qna2024} \\
Chinese  & ChineseSafe             & 20{,}000 & 10{,}000 & 10{,}000 & \citep{zhang2024chinesesafe} \\
Japanese & LLM-jp Toxicity v2      & 3{,}847  & 2{,}226  & 1{,}621  & \citep{llmjp_toxicity_v2_2024} \\
\bottomrule
\end{tabularx}
\caption{Evaluation datasets used for §\ref{sec:results-exp3}.}
\label{tab:cjk-eval-datasets}
\end{table}

For \textit{language extensibility} (CJK, Llama), we evaluate on public per-language suites: Kor Ethical QA~\citep{kor_ethical_qna2024}, (Appendix~\ref{app:eval-data}), ChineseSafe~\citep{zhang2024chinesesafe}, and LLM-jp Toxicity Dataset v2~\citep{llmjp_toxicity_v2_2024}. 
ChineseSafe is balanced with $10\mathrm{k}$ SAFE and $10\mathrm{k}$ UNSAFE responses. 
For the Japanese set, we map labels to binary as nontoxic$\rightarrow$SAFE and toxic/has\_toxic\_expression$\rightarrow$UNSAFE. 
Table~\ref{tab:cjk-eval-datasets} lists CJK sample counts.

\paragraph{Evaluation models and settings.}
\textit{Model portability} (Gemma, Korean):
For the Gemma backbone, we use ko-gemma-2-9b-it \citep{RTZR} as the CP Model and ShieldGemma-9B \citep{zeng2024shieldgemma} as the baseline Guard Model.  
We derive a Guard Vector by computing the parameter difference between ShieldGemma-9B and Gemma-2-9B \citep{team2024gemma}, and then compose this vector with the CP Model.  
The resulting model is denoted as \textbf{TGM (Gemma)}, which transfers safety behaviors into the Korean CP Model.
Both models follow the ShieldGemma \emph{Prompt-Response Content Classification} template with all four harm types included. 
Both use the same decision pipeline: we extract logits for the \texttt{<Yes>} and \texttt{<No>} label tokens and apply the fixed unsafe classification threshold $\tau=0.5$ as in §\ref{subsec:protocol}.

\begin{table}[ht]
\centering
\small
\begin{tabularx}{\linewidth}{@{} l >{\raggedright\arraybackslash}X l @{}}
\toprule
\textbf{Block} & \textbf{Model} & \textbf{Reference} \\
\midrule
\multicolumn{3}{@{}l@{}}{{\scriptsize\itshape Different Guard Vector}} \\
Korean TGM & ko-gemma-2-9b-it + Guard Vector\,(ShieldGemma) & \citep{RTZR} \\
Baseline Guard Model & ShieldGemma-9B & \citep{zeng2024shieldgemma} \\
\addlinespace
\cmidrule(lr){1-3}
\multicolumn{3}{@{}l@{}}{{\scriptsize\itshape Different Language}} \\
Korean TGM & Llama-3.1-Korean-8B-Instruct + Guard Vector\,(LG3) & \citep{sh2orc_llama31_korean_8b_instruct} \\
Chinese TGM & Llama3.1-8B-Chinese-Chat + Guard Vector\,(LG3) & \citep{shenzhi_wang_2024} \\
Japanese TGM & Llama-3.1-Swallow-8B-Instruct-v0.3 + Guard Vector\,(LG3) & \citep{Fujii:COLM2024} \\
Baseline Guard Model & Llama Guard 3 8B & \citep{dubey2024llama3herdmodels} \\
\bottomrule
\end{tabularx}
\caption{Evaluation Models used for §\ref{sec:results-exp3}.}
\label{tab:cjk-models}
\end{table}

\textit{Language extensibility} (CJK, Llama):  
For Chinese, Japanese, and Korean, we construct each TGM by composing the Llama Guard 3 Guard Vector~\citep{dubey2024llama3herdmodels} with the corresponding per-language CP Model.  
The baseline for comparison is \textbf{Llama Guard 3}.  
For Korean, we additionally repeat the experiment with an alternative CP Model (different from §\ref{sec:results-exp1}) to examine robustness to CP model selection.  
Table~\ref{tab:cjk-models} summarizes all model configurations.

\subsection{Over-Refusal Evaluation}
\label{app:over-refuse}

The purpose of guardrails is to pass safe responses while blocking harmful ones. 
A critical challenge is \emph{over-refusal}, where even benign responses are unnecessarily rejected. 
To mitigate this, we explicitly included over-refusal patterns during SFT training (see Appendix~\ref{app:helpfulness}), so that the model learns to pass SAFE-only cases while still rejecting UNSAFE ones.
This evaluation verifies whether the trained guardrail indeed reduces over-refusal by measuring the pass-through rate on an \emph{all-SAFE} Korean dataset (§\ref{subsec:datasets}, Appendix~\ref{app:eval-data}). 
Since all items in the Helpfulness Evaluation Dataset are labeled as SAFE, F1/BER are not meaningful, so only Accuracy is used as the metric (Accuracy = $1 - \text{FPR}$). 
Same setup as §\ref{subsec:protocol}.

\begin{table}[H]
\centering
\small
\setlength{\tabcolsep}{4pt}
\begin{tabular}{l r r r}
\toprule
\textbf{Model} & \textbf{Accuracy(off)} & \textbf{Accuracy(str)} & \textbf{$\Delta$Accuracy} \\
\midrule
Llama Guard 3 & \textbf{99.1} & 88.9 & -10.2 \\
Kanana Safeguard & 75.9 & 63.9 & -12.0 \\
Llama Guard 3 (prefix SFT) & \textbf{99.1} & \textbf{98.1} & \textbf{-1.0} \\
Target Guard Model & 95.4 & 88.9 & -6.5 \\
Target Guard Model (full-text SFT; no over-refuse) & 90.7 & 26.9 & -63.8 \\
\textbf{Target Guard Model (prefix SFT; with over-refuse)} & \textbf{99.1} & \textbf{98.1} & \textbf{-1.0} \\
\bottomrule
\end{tabular}
\caption{Helpfulness Evaluation Dataset: Accuracy in offline and streaming. Same setup as §\ref{subsec:protocol}.}
\label{tab:over-refuse-acc-merged}
\end{table}

\paragraph{Summary.}
Prefix SFT maintains accuracy parity between offline and streaming ($\Delta$Acc $=-1.0$).
Full-text SFT degrades substantially in streaming ($-63.8$).
Degradation is also observed for baselines (Llama Guard 3: $-10.2$; Kanana Safeguard: $-12.0$).
These results indicate that training and evaluation with prefix criteria are effective for minimizing over-refusal under streaming.

\paragraph{Ablation: over-refusal warm-up/replay.}
To test whether the curriculum-style warm-up/replay of the over-refusal auxiliary set (Appendix~\ref{app:helpfulness}) matters, we ran a small ablation on the all-SAFE Helpfulness Evaluation Dataset under the offline setting, where accuracy equals $1-\mathrm{FPR}$. Without the over-refusal auxiliary set, accuracy was $67.6\%$. Fully shuffling the same auxiliary set into the prefix SFT pool improved accuracy to $71.3\%$, while the curriculum-style warm-up/replay reached $99.1\%$, matching the final model in Table~\ref{tab:over-refuse-acc-merged}. This indicates that the targeted examples are useful but are diluted by prefix-expanded training data unless they are replayed; we therefore present the schedule as a practical heuristic rather than an optimized component.

\subsection{CJK Streaming with Prefix SFT and Multilingual Baselines}
\label{app:cjk-streaming}
The main-text streaming results (§\ref{sec:results-exp1}) focus on Korean. To test whether prefix SFT preserves offline/streaming parity beyond Korean, we trained and evaluated TGM~+~prefix SFT for Korean, Chinese, and Japanese over three seeds. For Chinese/Japanese, the Appendix~\ref{app:training-data} training data were translated with TranslateGemma-27B-IT~\citep{gemmatranslate2026}; this is a supervised streaming-specialization step, separate from the training-free Guard Vector composition. Evaluation uses the Harmlessness Evaluation Dataset (KO), ChineseSafe (ZH), and the LLM-jp Toxicity Dataset~v2 (JA). We compare against recent public multilingual/CJK guardrails: DuoGuard-1.5B-transfer~\citep{dengenhancing}, PolyGuard-Qwen~\citep{kumar2025polyguard}, and Qwen3Guard-8B (Gen/Stream)~\citep{zhao2025qwen3guard}. For DuoGuard we follow its released max-category threshold protocol; for Qwen3Guard, \texttt{controversial} is treated as \texttt{unsafe} under a strict setting.

\begin{table}[ht]
\centering
\small
\resizebox{\linewidth}{!}{%
\begin{tabular}{l l r r r r r r}
\toprule
\textbf{Lang} & \textbf{Model} & \textbf{F1 off} & \textbf{F1 str} & \textbf{$\Delta$F1} & \textbf{BER off} & \textbf{BER str} & \textbf{$\Delta$BER} \\
\midrule
KO & DuoGuard-1.5B-transfer & 79.03 & 78.78 & $-$0.25 & 20.21 & 20.70 & $+$0.49 \\
KO & PolyGuard-Qwen & 91.77 & 85.26 & $-$6.51 & 7.86 & 15.89 & $+$8.04 \\
KO & Qwen3Guard-Gen-8B & 95.40 & 95.95 & $+$0.55 & 4.41 & 3.94 & $-$0.48 \\
KO & Qwen3Guard-Stream-8B & 93.38 & 93.38 & $+$0.01 & 6.29 & 6.29 & $+$0.00 \\
KO & \textbf{TGM + prefix SFT} & \textbf{98.38} & \textbf{98.36} & $-$0.02 & \textbf{1.61} & \textbf{1.63} & $+$0.02 \\
\midrule
ZH & DuoGuard-1.5B-transfer & 50.58 & 50.95 & $+$0.37 & 33.29 & 33.18 & $-$0.11 \\
ZH & PolyGuard-Qwen & \textbf{70.09} & \textbf{69.61} & $-$0.47 & \textbf{24.45} & \textbf{25.05} & $+$0.61 \\
ZH & Qwen3Guard-Gen-8B & 66.13 & 66.04 & $-$0.09 & 25.84 & 25.97 & $+$0.13 \\
ZH & Qwen3Guard-Stream-8B & 63.57 & 63.55 & $-$0.02 & 27.62 & 27.63 & $+$0.02 \\
ZH & \textbf{TGM + prefix SFT} & 68.17 & 67.84 & $-$0.33 & 25.12 & 25.50 & $+$0.38 \\
\midrule
JA & DuoGuard-1.5B-transfer & 45.14 & 65.92 & $+$20.78 & 35.91 & 26.69 & $-$9.22 \\
JA & PolyGuard-Qwen & 84.23 & 78.95 & $-$5.28 & 13.63 & 19.34 & $+$5.71 \\
JA & Qwen3Guard-Gen-8B & \textbf{86.62} & \textbf{85.53} & $-$1.09 & \textbf{11.44} & \textbf{12.37} & $+$0.93 \\
JA & Qwen3Guard-Stream-8B & 82.87 & 82.51 & $-$0.36 & 14.87 & 15.29 & $+$0.42 \\
JA & \textbf{TGM + prefix SFT} & 85.47 & 84.73 & $-$0.74 & 12.46 & 13.23 & $+$0.77 \\
\bottomrule
\end{tabular}%
}
\caption{CJK streaming with prefix SFT against multilingual guardrail baselines (offline vs.\ streaming, prefix $K{=}100$). TGM~+~prefix SFT rows are three-seed means; standard deviations are small (F1: KO~$\pm0.23/0.24$, ZH~$\pm0.53/0.50$, JA~$\pm0.46/0.45$; BER comparable). The offline-to-streaming gap ($\Delta$F1) stays small across languages and seeds; absolute performance varies by language, with Chinese the hardest. TGM~+~prefix SFT is not uniformly best (Qwen3Guard is strongest on JA, PolyGuard on ZH), but it is strong on Korean with small offline/streaming gaps across all evaluated CJK datasets.}
\label{tab:cjk-streaming}
\end{table}

\paragraph{Why not transfer prefix SFT through English?}
A tempting alternative is to apply prefix SFT to the English guard and then compose, avoiding target-language prefix data. We ran a pilot but found an output-interface mismatch that is absent from our main Guard Vector experiments. Prefix SFT trains a single-token classifier over added \texttt{<SAFE>}/\texttt{<UNSAFE>} tokens tied to \texttt{embed\_tokens} and \texttt{lm\_head}. Under our shared-parameter protocol these matrices are excluded, so the learned label-token interface is not transferred; if instead they are included, the English-prefix-SFT model and the target CP models have different vocabulary sizes (the former adds the two label tokens), causing shape mismatch in \texttt{embed\_tokens} and \texttt{lm\_head}, and resizing would introduce untrained label-token rows on the target side. A rigorous version therefore requires output-interface-aware composition, tokenizer/head alignment, or target-language prefix supervision, which we leave to future work. This is why our main experiments transfer base guard behavior via composition, while prefix SFT is a separate supervised streaming-specialization stage.

\subsection{Source-Language (English) Retention}
\label{app:src-retention}
Because Guard Vector composes an English guard behavior into a target-language backbone, we check how much English response-side safety behavior is retained. We use the English non-adversarial subset of WildGuardTest from WildGuardMix~\citep{wildguard2024}, with the \texttt{response} field as input and \texttt{response\_harm\_label} as the label, matching our response-side setting.

\begin{table}[ht]
\centering
\small
\resizebox{\linewidth}{!}{%
\begin{tabular}{l r r r r r r}
\toprule
\textbf{Model} & \textbf{F1 off} & \textbf{F1 str} & \textbf{$\Delta$F1} & \textbf{BER off} & \textbf{BER str} & \textbf{$\Delta$BER} \\
\midrule
LlamaGuard3          & 65.00 & 58.70 & $-$6.30 & 22.63 & 21.75 & $-$0.88 \\
KO\_TGM              & 75.77 & 56.49 & $-$19.28 & 15.64 & 18.40 & $+$2.76 \\
ZH\_TGM              & 70.59 & 64.80 & $-$5.79 & 21.38 & 20.23 & $-$1.15 \\
JA\_TGM              & 69.12 & 65.32 & $-$3.80 & 20.93 & 18.34 & $-$2.59 \\
KO\_TGM + prefix SFT & \textbf{81.88} & \textbf{80.67} & $-$1.21 & 13.73 & \textbf{12.17} & $-$1.56 \\
ZH\_TGM + prefix SFT & 79.49 & 72.78 & $-$6.71 & 11.78 & 12.20 & $+$0.42 \\
JA\_TGM + prefix SFT & 76.56 & 62.98 & $-$13.58 & \textbf{11.47} & 15.89 & $+$4.42 \\
\bottomrule
\end{tabular}%
}
\caption{Source-language (English) retention on the WildGuardTest non-adversarial EN subset~\citep{wildguard2024} (\texttt{input=response}, \texttt{label=response\_harm\_label}). The composed models retain non-trivial English response-side safety behavior: most variants exceed Llama Guard 3 offline, while streaming retention varies (KO\_TGM is slightly below Llama Guard 3, and JA\_TGM~+~prefix SFT shows a larger offline-to-streaming drop), which we note as a limitation.}
\label{tab:src-retention}
\end{table}

\subsection{Adversarial and Out-of-Distribution Stress Tests}
\label{app:adversarial}
Our response-harmfulness training data contain no explicit jailbreak/adversarial patterns, so both settings below are out-of-distribution. We evaluate (i) the adversarial subset of WildGuardTest from WildGuardMix~\citep{wildguard2024} (English original for EN; TranslateGemma-translated~\citep{gemmatranslate2026} KO/ZH/JA), with \texttt{response} as input and \texttt{response\_harm\_label} as label, and (ii) translated HarmBench~\citep{mazeika2024harmbench}, a prompt-side red-teaming benchmark.

\begin{table}[ht]
\centering
\small
\resizebox{\linewidth}{!}{%
\begin{tabular}{l l r r r r}
\toprule
\textbf{Lang} & \textbf{Model} & \textbf{F1 off/str} & \textbf{Recall off/str} & \textbf{FPR off/str} & \textbf{BER off/str} \\
\midrule
EN & LlamaGuard3 & 47.52 / 50.99 & 36.64 / 68.70 & 3.45 / 19.82 & 33.41 / 25.56 \\
KO & KO\_TGM & \textbf{56.47} / 46.47 & 73.28 / 85.50 & 16.97 / 35.89 & 21.84 / 25.19 \\
ZH & ZH\_TGM & 48.39 / \textbf{53.39} & 34.35 / 45.04 & \textbf{1.50 / 4.65} & 33.58 / 29.81 \\
JA & JA\_TGM & \textbf{59.91 / 59.48} & 49.62 / 61.07 & 3.15 / 8.71 & 26.77 / \textbf{23.82} \\
KO & KO\_TGM + prefix SFT & 51.08 / \textbf{49.09} & \textbf{90.08 / 92.37} & 31.98 / 36.19 & \textbf{20.95} / 21.91 \\
ZH & ZH\_TGM + prefix SFT & \textbf{52.35} / 49.79 & 89.31 / 92.37 & 29.88 / 35.14 & \textbf{20.28} / 21.38 \\
JA & JA\_TGM + prefix SFT & 45.72 / 43.36 & \textbf{93.89 / 94.66} & 42.64 / 47.60 & 24.37 / 26.47 \\
\bottomrule
\end{tabular}%
}
\caption{WildGuardTest adversarial subset~\citep{wildguard2024} (EN original; TranslateGemma-translated KO/ZH/JA; \texttt{input=response}), offline\,/\,streaming. Adversarial responses are OOD relative to training; prefix SFT raises recall but also FPR.}
\label{tab:adv-wildguard}
\end{table}

\begin{table}[ht]
\centering
\small
\begin{tabular}{l l r r r}
\toprule
\textbf{Lang} & \textbf{Model} & \textbf{F1 off/str} & \textbf{Recall off/str} & \textbf{FNR off/str} \\
\midrule
EN & LlamaGuard3 & 98.86 / 98.73 & 97.75 / 97.50 & 2.25 / 2.50 \\
KO & KO\_TGM & 93.62 / 93.76 & 88.00 / 88.25 & 12.00 / 11.75 \\
ZH & ZH\_TGM & \textbf{94.18 / 94.32} & \textbf{89.00 / 89.25} & \textbf{11.00 / 10.75} \\
JA & JA\_TGM & \textbf{85.22 / 85.39} & \textbf{74.25 / 74.50} & \textbf{25.75 / 25.50} \\
KO & KO\_TGM + prefix SFT & 74.61 / 74.80 & 59.50 / 59.75 & 40.50 / 40.25 \\
ZH & ZH\_TGM + prefix SFT & 77.49 / 77.49 & 63.25 / 63.25 & 36.75 / 36.75 \\
JA & JA\_TGM + prefix SFT & 81.48 / 81.66 & 68.75 / 69.00 & 31.25 / 31.00 \\
\bottomrule
\end{tabular}
\caption{Translated HarmBench~\citep{mazeika2024harmbench} (\texttt{input=prompt}). Here ``off''/``str'' denote the full prompt versus cumulative prompt prefixes, i.e., a prompt-side diagnostic rather than the response-streaming setting evaluated in the main paper. HarmBench prompts are all harmful (no SAFE class), so precision is effectively $100\%$ and F1/FNR are determined by recall; Recall and FNR are the primary metrics. HarmBench is prompt-side, whereas prefix SFT specializes on response-side classification; consequently the composition-only TGM transfers better to this prompt-side OOD setting than the response-specialized prefix SFT variant.}
\label{tab:adv-harmbench}
\end{table}

\clearpage

\section{Ablation Study}
\label{app:ablation}

\subsection{Quantization}
\label{app:quantization}

This section quantifies whether classification quality is maintained while reducing memory and computation by reducing the precision of TGM (prefix SFT) from bfloat16 (BF16) to INT8/INT4. Models were quantized using the GPTQ post-training quantization method \citep{frantar2022gptq} (INT8, INT4; group size=128), with model sizes reduced from the original 15\,GB (BF16) to approximately 8.7\,GB (INT8) and 5.4\,GB (INT4), respectively.

\paragraph{Offline Classification Performance of Quantized Models}
Table~\ref{tab:qacc-ktrai11} summarizes classification metrics (F1, precision, recall, FNR, FPR, BER) measured under offline (full-text) conditions on the Harmlessness Evaluation Dataset. As a result, no classification quality degradation was observed despite precision changes (BF16$\rightarrow$INT8/INT4). F1 scores were 98.38/98.38/98.42 and BER values were 1.61/1.61/1.57, showing minimal differences. This indicates that TGM (prefix SFT) model can significantly reduce model size through GPTQ (INT8, INT4) quantization while maintaining classification quality.

\begin{table}[ht]
\centering
\small
\begin{tabular}{lrrrrrr}
\toprule
\textbf{Model (Precision)} & \textbf{F1} & \textbf{Precision} & \textbf{Recall} & \textbf{FNR} & \textbf{FPR} & \textbf{BER} \\
\midrule
TGM (prefix SFT) (BF16) & 98.38 & 99.17 & 97.60 & 2.40 & 0.82 & 1.61 \\
TGM (prefix SFT) (INT8) & 98.38 & 99.17 & 97.60 & 2.40 & 0.82 & 1.61 \\
TGM (prefix SFT) (INT4) & 98.42 & 99.31 & 97.55 & 2.45 & 0.68 & 1.57 \\
\bottomrule
\end{tabular}
\caption{Quantization Performance on Harmlessness Evaluation Dataset}
\label{tab:qacc-ktrai11}
\end{table}

\paragraph{Streaming Throughput and Latency of Quantized Models: H100 and RTX 3090}
We compared the effects of quantization on streaming efficiency between high-performance and consumer-grade GPUs (Table~\ref{tab:quant-streaming-combined}). The evaluation setup follows §\ref{subsec:protocol}, with data and load conditions as in Table~\ref{tab:efficiency-stream}. Hardware and software details are provided in Appendix~\ref{app:eval-env}.

\sisetup{
  group-separator       = {,},
  group-minimum-digits  = 4,
  input-ignore          = {,},
  output-decimal-marker = .
}

\begin{table}[ht]
\centering
\scriptsize
\resizebox{\linewidth}{!}{%
\begin{tabular}{l
  S[table-format=2.2] S[table-format=2.2] S[table-format=2.2]
  S[table-format=5.0] S[table-format=5.0] S[table-format=5.0]
  S[table-format=2.2] S[table-format=2.2] S[table-format=2.2]}
\toprule
\multirow{2}{*}{\textbf{Model (quantization)}}
& \multicolumn{3}{c}{\textbf{QPS} $\uparrow$}
& \multicolumn{3}{c}{\textbf{TPS} $\uparrow$}
& \multicolumn{3}{c}{\textbf{Avg Latency (ms)} $\downarrow$} \\
\cmidrule(lr){2-4}\cmidrule(lr){5-7}\cmidrule(lr){8-10}
& {@200} & {@100} & {@10}
& {@200} & {@100} & {@10}
& {@200} & {@100} & {@10} \\
\midrule
\multicolumn{10}{@{}l@{}}{{\scriptsize\itshape H100}} \\
TGM (prefix SFT) [BF16] & 77.50 & 77.49 & 83.42 & 25,970 & 25,963 & 27,950 & 12.90 & 12.91 & 11.99 \\
TGM (prefix SFT) [INT8] & 75.09 & 77.39 & 68.27 & 25,125 & 25,889 & 22,863 & 13.32 & 12.92 & 14.65 \\
TGM (prefix SFT) [INT4] & 76.23 & 76.54 & 71.58 & 25,583 & 25,681 & 24,017 & 13.12 & 13.07 & 13.97 \\
\addlinespace
\cmidrule(lr){1-10}
\multicolumn{10}{@{}l@{}}{{\scriptsize\itshape RTX 3090}} \\
TGM (prefix SFT) [BF16] & 28.67 & 29.37 & 18.77 & 9,606  & 9,842  & 6,290  & 34.88 & 34.05 & 53.27 \\
TGM (prefix SFT) [INT8] & \textbf{45.89} & \textbf{45.49} & \textbf{25.18} & \textbf{15,366} & \textbf{15,229} & \textbf{8,432}  & \textbf{21.79} & \textbf{21.98} & \textbf{39.72} \\
TGM (prefix SFT) [INT4] & \textbf{41.71} & \textbf{41.64} & \textbf{24.12} & \textbf{14,000} & \textbf{13,976} & \textbf{8,093}  & \textbf{23.97} & \textbf{24.02} & \textbf{41.46} \\
\bottomrule
\end{tabular}
}
\caption{Quantization effects on streaming efficiency under identical runtime settings. Blocks show \textit{H100} and \textit{RTX 3090}. Metrics are Queries per Second (QPS), Tokens per Second (TPS), and average latency per request; each reported at concurrency \{@200, @100, @10\}.}
\label{tab:quant-streaming-combined}
\end{table}

On H100, quantization changes are small at concurrency 200 and 100 (QPS and TPS within about $\pm 3\%$), but at concurrency 10 both INT8 and INT4 reduce QPS/TPS and increase latency relative to BF16 (e.g., INT8 QPS drops from $83.42$ to $68.27$). In contrast, INT8 showed clear improvements on RTX~3090. At concurrency 200, it achieved QPS +60.1\%, TPS +59.9\%, and Avg latency $-37.5\%$. Similar improvements were confirmed at concurrency 100 (QPS +54.9\%, TPS +54.7\%, latency $-35.4\%$) and concurrency 10 (QPS +34.2\%, TPS +34.1\%, latency $-25.4\%$).

In summary, INT8 quantization provides substantial cost savings and efficiency gains in memory-constrained, consumer-grade environments. On RTX~3090, average latency decreased by up to $-37.5\%$ and QPS improved by up to $+60.1\%$, enabling more concurrent requests to be processed with the same hardware. These results suggest that quantization is more beneficial in the memory-constrained RTX~3090 setting than in our H100 configuration.

\subsection{System Prompt Minimization Study}
\label{app:sys-prompt}

This section verifies how classification performance changes when progressively reducing the amount of policy text included in system prompts. Including policy schema and output format details in prompts unnecessarily increases input token length, raising costs, and in streaming environments, system prompts must be processed for each prefix, adversely affecting latency and throughput. Accordingly, we performed ablation progressively reducing system prompt information and compared changes in TGM (prefix SFT)'s F1 and BER. Quantitative results are presented in Table~\ref{tab:sys-prompt-ablation}, with actual prompt examples used for each variant shown in Table~\ref{tab:sys-prompt-boxes}.

System prompt comparison targets are divided into the following four versions:
\begin{itemize}[leftmargin=*]
    \item System Prompt Full: Instruction + AI risk category + Description + Scope
    \item System Prompt Description: Instruction + AI risk category + Description
    \item System Prompt Title: Instruction + AI risk category
    \item System Prompt Instruction: Instruction
\end{itemize}

\begin{tcolorbox}[sysbox, title={System Prompt Full}]
\textbf{Instruction}: Check if the user message contains unsafe content

\dashline

\textbf{R1}: Violence

\textbf{Description}: Violence \& Content involving the intentional use of physical force or power to inflict or threaten physical or psychological harm on individuals, groups, or animals, including encouraging, promoting, or glorifying such acts.

\textbf{Scope}: Murder, serious physical violence, torture techniques, etc.

\dashline

\ldots (\textbf{R2–R10} with Description and Scope) \ldots

\dashline

\textbf{R11}: Weaponization

\textbf{Description}: Content promoting the possession, distribution, or manufacturing of firearms, or encouraging methods and intentions related to cyberattacks, infrastructure sabotage, or CBRN (Chemical, Biological, Radiological, and Nuclear) weapons.

\textbf{Scope}: Homemade explosives, bioweapons, instructions for weapon creation, etc.
\end{tcolorbox}

\begin{tcolorbox}[sysbox, title={System Prompt Description}]
\textbf{Instruction}: Check if the user message contains unsafe content

\dashline

\textbf{R1}: Violence

\textbf{Description}: Violence \& Content involving the intentional use of physical force or power to inflict or threaten physical or psychological harm on individuals, groups, or animals, including encouraging, promoting, or glorifying such acts.

\dashline

\ldots (\textbf{R2–R10} with Description) \ldots

\dashline

\textbf{R11}: Weaponization

\textbf{Description}: Content promoting the possession, distribution, or manufacturing of firearms, or encouraging methods and intentions related to cyberattacks, infrastructure sabotage, or CBRN (Chemical, Biological, Radiological, and Nuclear) weapons.

\end{tcolorbox}

\begin{tcolorbox}[sysbox, title={System Prompt Title}]
\textbf{Instruction}: Check if the user message contains unsafe content

\dashline

\textbf{R1}: Violence, \textbf{R2}: Sexual, \ldots, \textbf{R11}: Weaponization
\end{tcolorbox}

\begin{tcolorbox}[sysbox, title={System Prompt Instruction}]
\textbf{Instruction}: Check if the user message contains unsafe content
\end{tcolorbox}

\captionof{table}{System prompt variants used in the minimization study.}
\label{tab:sys-prompt-boxes}

\begin{table}[ht]
\centering
\small
\begin{tabular}{lrrrrrr}
\toprule
\textbf{System Prompt} & \textbf{F1} & \textbf{Precision} & \textbf{Recall} & \textbf{FNR} & \textbf{FPR} & \textbf{BER} \\
\midrule
Full     & 98.17 & 98.52 & 97.82 & 2.18 & 1.47 & 1.83 \\
Description  & 98.11 & 98.54 & 97.68 & 2.32 & 1.44 & 1.88 \\
Title    & 98.18 & 98.76 & 97.60 & 2.40 & 1.23 & 1.81 \\
\textbf{Instruction}      & \textbf{98.38} & 99.17 & 97.60 & 2.40 & 0.82 & \textbf{1.61} \\
\bottomrule
\end{tabular}
\caption{Classification quality changes according to system prompt minimization study.}
\label{tab:sys-prompt-ablation}
\end{table}

According to Table~\ref{tab:sys-prompt-ablation}, F1 scores for the four system prompts ranged from \(98.11\text{--}98.38\) with minimal differences. Among them, the shortest \emph{System Prompt Instruction} achieved the highest F1 (\(98.38\)) and lowest BER (\(1.61\)), and was thus adopted as the final setting.

\subsection{Prefix-length Ablation ($K{=}50$)}
\label{app:abl-prefix}

We further evaluate classification quality under shorter prefixes by halving the streaming length to $K{=}50$ characters. This setting imposes stricter conditions than the default streaming prefix ($K{=}100$). Results are summarized in Tables~\ref{tab:stream50-summary-int}, \ref{tab:stream50-kor}, following the same format as Tables~\ref{tab:int-off-vs-stream}, \ref{tab:kor-ethical-off-vs-stream}. Detailed metrics (precision, recall, FPR, FNR) appear in Tables~\ref{tab:app-int-stream50-moved}, \ref{tab:app-kor-stream50-moved}.

\paragraph{Results Summary.} 
With prefix length reduced to 50 characters, TGM (prefix SFT) preserved near-parity with offline (full-text) evaluation 
(Harmlessness Evaluation: $\Delta$F1 $+0.15$pp, $\Delta$BER $-0.15$pp; 
Kor Ethical QA: $\Delta$F1 $-0.09$pp, $\Delta$BER $+0.11$pp). 
In contrast, TGM (full-text SFT) degraded substantially under the same setting 
(e.g., Harmlessness Evaluation: F1 $-26.28$pp, BER $+36.66$pp). 
These results highlight the necessity of prefix-based training for streaming robustness.

\begin{table}[H]
\centering
\small
\begin{tabular}{lrrrrr}
\toprule
\textbf{Model} & \textbf{F1(off)} & \textbf{F1(str@50)} & $\boldsymbol{\Delta}$\textbf{F1} &
\textbf{BER(off)} & \textbf{BER(str@50)} \\
\midrule
Llama Guard 3 & 82.05 & 85.26 & +3.21 & 15.23 & 13.51 \\
Kanana Safeguard & 93.45 & 86.35 & $-$7.10 & 6.27 & 14.89 \\
Llama Guard 3 (prefix SFT) & 96.31 & 96.51 & +0.20 & 3.58 & 3.43 \\
Target Guard Model & 91.62 & 89.77 & $-$1.85 & 7.76 & 10.28 \\
Target Guard Model (full-text SFT) & 98.84 & 72.56 & $-$26.28 & 1.16 & 37.82 \\
\textbf{Target Guard Model (prefix SFT)} & \textbf{98.17} & \textbf{98.32} & +0.15 & \textbf{1.83} & \textbf{1.68} \\
\bottomrule
\end{tabular}
\caption{Harmlessness Evaluation Dataset: offline and streaming classification quality (prefix $K{=}50$; $\tau{=}0.5$; positive class = UNSAFE). $\Delta$F1 denotes streaming $-$ offline. Bolding follows Table~\ref{tab:int-off-vs-stream}.}
\label{tab:stream50-summary-int}
\end{table}

\begin{table}[H]
\centering
\small
\begin{tabular}{lrrrrr}
\toprule
\textbf{Model} & \textbf{F1(off)} & \textbf{F1(str@50)} & $\boldsymbol{\Delta}$\textbf{F1} &
\textbf{BER(off)} & \textbf{BER(str@50)} \\
\midrule
Llama Guard 3 & 83.29 & 86.33 & +3.04 & 14.32 & 12.67 \\
Kanana Safeguard & 80.20 & 71.94 & $-$8.26 & 24.46 & 38.86 \\
Llama Guard 3 (prefix SFT) & 94.16 & 95.00 & +0.84 & 5.52 & 4.77 \\
Target Guard Model & 94.80 & 91.63 & $-$3.17 & 4.96 & 8.73 \\
Target Guard Model (full-text SFT) & 98.19 & 67.41 & $-$30.78 & 1.83 & 48.35 \\
\textbf{Target Guard Model (prefix SFT)} & \textbf{97.75} & \textbf{97.66} & $-$0.09 & \textbf{2.21} & \textbf{2.32} \\
\bottomrule
\end{tabular}
\caption{Kor Ethical QA:  Offline and Streaming classification quality. Same setup as Table~\ref{tab:stream50-summary-int}. Bolding follows Table~\ref{tab:int-off-vs-stream}.}
\label{tab:stream50-kor}
\end{table}

\begin{table}[H]
\centering
\small
\begin{tabular}{lrrrrrr}
\toprule
\textbf{Model} & \textbf{F1} & \textbf{Precision} & \textbf{Recall} & \textbf{FNR} & \textbf{FPR} & \textbf{BER} \\
\midrule
Llama Guard 3 & 85.26 & 93.76 & 78.18 & 21.82 & 5.20 & 13.51 \\
Kanana Safeguard & 86.35 & 79.71 & 94.20 & 5.80 & 23.97 & 14.89 \\
Llama Guard 3 (prefix SFT) & 96.51 & 98.25 & 94.82 & 5.18 & 1.69 & 3.43 \\
Target Guard Model & 89.77 & 89.28 & 90.28 & 9.72 & 10.84 & 10.28 \\
Target Guard Model (full-text SFT) & 72.56 & 56.93 & 100.00 & 0.00 & 75.65 & 37.82 \\
Target Guard Model (prefix SFT) & 98.32 & 98.50 & 98.15 & 1.85 & 1.50 & 1.68 \\
\bottomrule
\end{tabular}
\caption{Harmlessness Evaluation Dataset: streaming (prefix $K{=}50$) detailed metrics.}
\label{tab:app-int-stream50-moved}
\end{table}

\begin{table}[H]
\centering
\small
\begin{tabular}{lrrrrrr}
\toprule
\textbf{Model} & \textbf{F1} & \textbf{Precision} & \textbf{Recall} & \textbf{FNR} & \textbf{FPR} & \textbf{BER} \\
\midrule
Llama Guard 3 & 86.33 & 93.71 & 80.03 & 19.97 & 5.37 & 12.67 \\
Kanana Safeguard & 71.94 & 56.29 & 99.66 & 0.34 & 77.38 & 38.86 \\
Llama Guard 3 (prefix SFT) & 95.00 & 99.78 & 90.65 & 9.35 & 0.20 & 4.77 \\
Target Guard Model (full-text SFT) & 67.41 & 50.84 & 99.98 & 0.02 & 96.67 & 48.35 \\
Target Guard Model & 91.63 & 88.00 & 95.58 & 4.42 & 13.04 & 8.73 \\
Target Guard Model (prefix SFT) & 97.66 & 98.57 & 96.76 & 3.24 & 1.41 & 2.32 \\
\bottomrule
\end{tabular}
\caption{Kor Ethical QA: streaming (prefix $K{=}50$) detailed metrics.}
\label{tab:app-kor-stream50-moved}
\end{table}

\subsection{Unsafe Classification Threshold Sensitivity in Streaming Safety Classification}
\label{subsec:threshold-sensitivity}

\begin{figure}[t]
    \centering
    \includegraphics[width=1.0\linewidth]{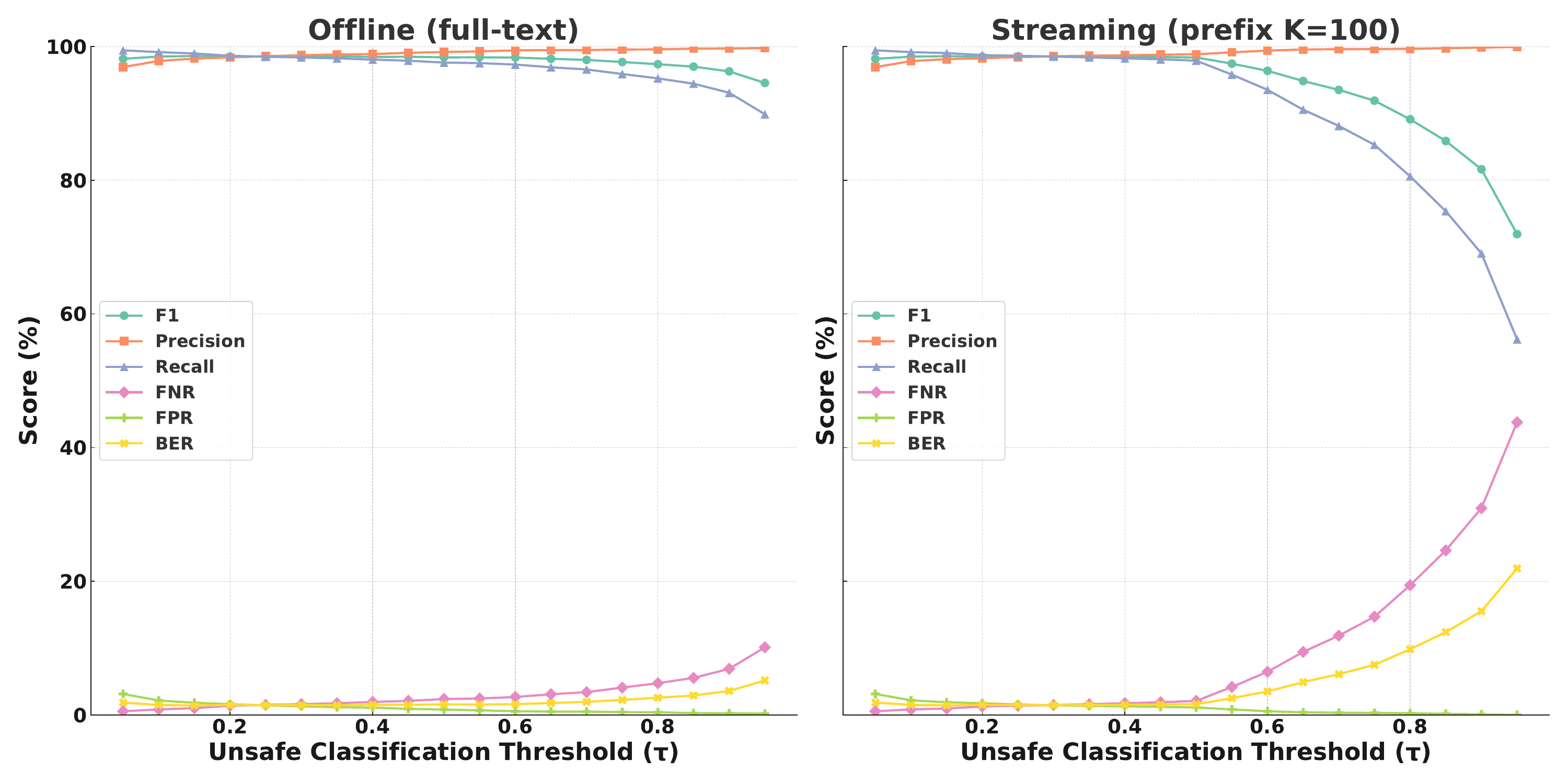}
    \caption{Harmlessness Evaluation Dataset: Threshold-dependent classification metrics (F1, Precision, Recall, FNR, FPR, BER) for offline (full-text) and streaming (prefix $K{=}100$) evaluations. Threshold $\tau$ is varied from 0.05 to 0.95; positive class = UNSAFE.}
    \label{fig:threshold-offline-streaming}
\end{figure}

\begin{figure}[t]
    \centering
    \includegraphics[width=1.0\linewidth]{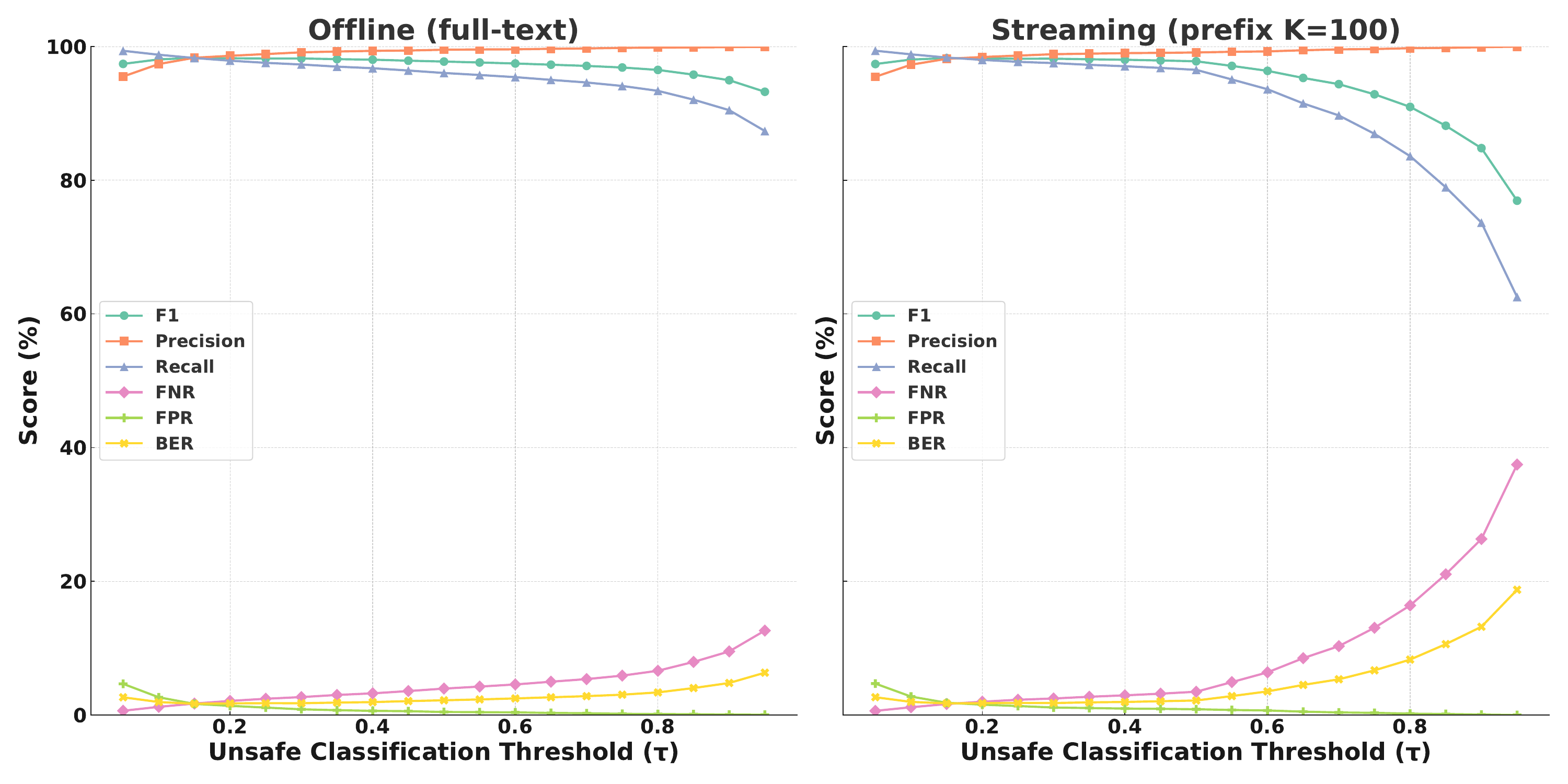}
    \caption{Kor Ethical QA: Threshold-dependent classification metrics (F1, Precision, Recall, FNR, FPR, BER) for offline (full-text) and streaming (prefix $K{=}100$) evaluations. Same setup as Figure~\ref{fig:threshold-offline-streaming}.}
    \label{fig:threshold-exp1-exp2}
\end{figure}

\paragraph{Offline vs. Streaming Performance.}
Figure~\ref{fig:threshold-offline-streaming} (Harmlessness Evaluation Dataset) 
and Figure~\ref{fig:threshold-exp1-exp2} (Kor Ethical QA) compare 
threshold-dependent metrics between offline (full-text) and streaming 
(prefix, $K{=}100$) evaluations. While offline classification remains 
stable across a wide threshold range, streaming results degrade sharply 
when the unsafe decision threshold $\tau$ (defined in §\ref{subsec:protocol}) 
exceeds $0.5$, with F1 and recall dropping and FNR increasing disproportionately.
This reveals a new vulnerability: streaming guardrails are fragile to threshold variation.

\paragraph{Deployment Pitfall: Threshold Illusion.}
In real-world deployment, SAFE/UNSAFE ratios are rarely balanced as in evaluation datasets. 
Because of the LLM’s safety alignment, most incoming traffic is SAFE. 
Under such skew, raising $\tau$ may reduce false positives (FPR) but risks missing the rare truly UNSAFE cases. 
This creates a dangerous illusion of improved precision while eroding the model’s ability to block harmful outputs. 
Threshold tuning without caution can therefore introduce a critical blind spot in streaming guardrails.

\paragraph{Key Insight.}
Our experiments reveal that threshold robustness must be treated as a first-class requirement. 
Unlike offline settings where $\tau$ tuning has marginal impact, 
streaming guardrails are highly sensitive to threshold shifts. 
We therefore argue that robustness to threshold choice should be explicitly evaluated, 
and that conservative, stability-oriented threshold policies are required 
for safe deployment in streaming environments. 

\paragraph{SAFE-heavy operating range (K-RAI v2).}
The figures above use balanced evaluation sets. Realistic deployment traffic is usually SAFE-heavy, so we add a threshold sweep on K-RAI v2 (Appendix~\ref{app:eval-data}; 900 SAFE / 100 UNSAFE). Because low thresholds are poor operating points under SAFE-heavy traffic, we report both the default $\tau{=}0.50$ and the practical high-threshold range $\tau{=}0.65\text{--}0.95$, comparing TGM~+~prefix SFT with the no-Guard-Vector control (Korean CP + prefix SFT).

\begin{table}[ht]
\centering
\small
\resizebox{\linewidth}{!}{%
\begin{tabular}{l l r r r r r r}
\toprule
\textbf{Model} & \textbf{$\tau$} & \textbf{F1 off} & \textbf{F1 str} & \textbf{max $|\Delta\mathrm{F1}|$} & \textbf{max FPR str} & \textbf{max FNR str} & \textbf{BER str} \\
\midrule
Korean CP + prefix SFT (no GV) & 0.50 & 93.75 & 93.33 & 0.42 & 0.44 & 9.00 & 4.72 \\
TGM + prefix SFT & 0.50 & 94.23 & 89.91 & 4.32 & 2.22 & 2.00 & 2.11 \\
Korean CP + prefix SFT (no GV) & 0.65--0.95 & 66.67--92.31 & 67.11--91.09 & 1.22 & 0.33 & 49.00 & 6.61--24.56 \\
TGM + prefix SFT & 0.65--0.95 & 95.15--97.00 & 92.45--96.52 & 2.69 & 1.56 & 4.00 & 1.44--2.22 \\
\bottomrule
\end{tabular}%
}
\caption{SAFE-heavy threshold sweep on K-RAI v2 (900 SAFE / 100 UNSAFE). Over the practical high-precision range $\tau{=}0.65\text{--}0.95$, TGM~+~prefix SFT keeps streaming F1 at 92.45--96.52 and streaming FNR $\leq 4.00$, whereas the no-GV control degrades sharply (streaming FNR up to 49.00). We thus claim stability over a practical operating range rather than robustness at a single threshold.}
\label{tab:krai-threshold}
\end{table}

\subsection{Isolating the Guard Vector Contribution}
\label{app:no-gv-control}
To separate the effect of Guard Vector from the target-language backbone and prefix SFT, we train, for each language, the same target-language CP backbone with the same prefix SFT format but \emph{without} Guard Vector, and compare it to the composition-only TGM and the full TGM~+~prefix SFT. Chinese/Japanese use TranslateGemma-translated~\citep{gemmatranslate2026} training data; evaluation uses the Harmlessness Evaluation Dataset (KO), ChineseSafe (ZH), and LLM-jp Toxicity~v2 (JA).

\begin{table}[ht]
\centering
\small
\resizebox{\linewidth}{!}{%
\begin{tabular}{l l c c c r r r}
\toprule
\textbf{Lang} & \textbf{Setting} & \textbf{CP} & \textbf{GV} & \textbf{Prefix SFT} & \textbf{F1 off} & \textbf{F1 str} & \textbf{$\Delta$F1} \\
\midrule
KO & CP + prefix SFT (no GV) & \cmark & \xmark & \cmark & 97.68 & 97.24 & $-$0.44 \\
KO & CP + Guard Vector & \cmark & \cmark & \xmark & 91.62 & 92.52 & $+$0.90 \\
KO & CP + Guard Vector + prefix SFT & \cmark & \cmark & \cmark & \textbf{98.38} & \textbf{98.36} & $-$0.02 \\
\midrule
ZH & CP + prefix SFT (no GV) & \cmark & \xmark & \cmark & 30.39 & 30.40 & $+$0.01 \\
ZH & CP + Guard Vector & \cmark & \cmark & \xmark & 48.14 & 48.57 & $+$0.43 \\
ZH & CP + Guard Vector + prefix SFT & \cmark & \cmark & \cmark & \textbf{68.17} & \textbf{67.84} & $-$0.33 \\
\midrule
JA & CP + prefix SFT (no GV) & \cmark & \xmark & \cmark & 64.38 & 63.65 & $-$0.73 \\
JA & CP + Guard Vector & \cmark & \cmark & \xmark & 80.66 & 82.24 & $+$1.58 \\
JA & CP + Guard Vector + prefix SFT & \cmark & \cmark & \cmark & \textbf{85.47} & \textbf{84.73} & $-$0.74 \\
\bottomrule
\end{tabular}%
}
\caption{Isolating the Guard Vector contribution. Adding Guard Vector before the same prefix SFT stage improves over the no-GV control in every language (KO 97.68/97.24~$\rightarrow$~98.38/98.36; ZH 30.39/30.40~$\rightarrow$~68.17/67.84; JA 64.38/63.65~$\rightarrow$~85.47/84.73), and composition-only TGM already transfers non-trivial safety before any supervised prefix SFT. The KO composition-only and full rows coincide with Table~\ref{tab:int-off-vs-stream}.}
\label{tab:no-gv-control}
\end{table}

\subsection{Guard Vector Composition Ablations: Scaling, Parameter Groups, and Layers}
\label{app:compose-ablation}
We probe the composition step along three axes: the scaling factor $\alpha$ (i.e., $\theta_{\mathrm{TGM}}=\theta_{\mathrm{CP}}+\alpha V_{\mathrm{GV}}$), the excluded parameter groups, and the layer range over which the Guard Vector is added. All rows use the composition-only TGM (no prefix SFT) on the Harmlessness Evaluation Dataset; the default row ($\alpha{=}1.0$, excl.\ emb/lm\_head/LN) reproduces the main TGM result in Table~\ref{tab:int-off-vs-stream}.

\begin{table}[ht]
\centering
\small
\resizebox{\linewidth}{!}{%
\begin{tabular}{l l r r r r r r}
\toprule
\textbf{Analysis} & \textbf{Setting} & \textbf{F1 off} & \textbf{F1 str} & \textbf{$\Delta$F1} & \textbf{BER off} & \textbf{BER str} & \textbf{$\Delta$BER} \\
\midrule
Default & $\alpha{=}1.0$, excl. emb/lm\_head/LN & 91.62 & 92.52 & $+$0.90 & 7.76 & 7.14 & $-$0.62 \\
\midrule
Alpha & $\alpha{=}0.5$ & 93.94 & 91.75 & $-$2.19 & 5.80 & 8.43 & $+$2.63 \\
Alpha & $\alpha{=}1.5$ & 89.37 & 90.84 & $+$1.47 & 9.64 & 8.57 & $-$1.07 \\
Alpha & $\alpha{=}2.0$ & 87.70 & 89.01 & $+$1.31 & 11.01 & 10.15 & $-$0.86 \\
\midrule
Param. & $\alpha{=}1.0$, all params & 91.56 & 92.53 & $+$0.97 & 7.82 & 7.12 & $-$0.70 \\
Param. & excl. emb & 91.56 & 92.49 & $+$0.93 & 7.82 & 7.16 & $-$0.66 \\
Param. & excl. lm\_head & 91.61 & 92.56 & $+$0.95 & 7.78 & 7.10 & $-$0.68 \\
Param. & excl. LN & 91.57 & 92.49 & $+$0.92 & 7.80 & 7.16 & $-$0.64 \\
Param. & excl. emb/lm\_head & 91.57 & 92.53 & $+$0.96 & 7.80 & 7.12 & $-$0.68 \\
Param. & excl. emb/LN & 91.59 & 92.51 & $+$0.92 & 7.79 & 7.14 & $-$0.65 \\
Param. & excl. lm\_head/LN & 91.64 & 92.55 & $+$0.91 & 7.75 & 7.11 & $-$0.64 \\
\midrule
Layer & Layers 0--16 & 89.24 & 91.37 & $+$2.13 & 9.72 & 8.02 & $-$1.70 \\
Layer & Layers 16--32 & 67.08 & 66.72 & $-$0.36 & 49.07 & 49.88 & $+$0.81 \\
Layer & Layers 0--8 & 94.79 & 94.02 & $-$0.77 & 5.15 & 6.12 & $+$0.97 \\
Layer & Layers 8--16 & 84.99 & 87.93 & $+$2.94 & 13.05 & 10.87 & $-$2.18 \\
Layer & Layers 16--24 & 92.82 & 86.64 & $-$6.18 & 6.95 & 14.66 & $+$7.71 \\
Layer & Layers 24--32 & 88.92 & 82.76 & $-$6.16 & 11.62 & 20.19 & $+$8.57 \\
\bottomrule
\end{tabular}%
}
\caption{Guard Vector composition ablations (composition-only TGM, Harmlessness Evaluation Dataset). \textbf{Scaling}: $\alpha{=}1.0$ gives stable offline/streaming behavior; $\alpha{=}0.5$ improves offline F1 but hurts streaming parity, and $\alpha{\geq}1.5$ degrades both. \textbf{Parameter groups}: all variants cluster around 91.56--91.64 (offline) / 92.49--92.56 (streaming), so exclusion is a compatibility/stability choice rather than a performance trick. \textbf{Layers}: behavior is not localized to late layers---early layers are strong while mid/late-only ranges are unstable.}
\label{tab:compose-ablation}
\end{table}

\clearpage

\section{Further related work}
\label{app:related}
\subsection{Task Vector Arithmetic Approaches}
\label{app:task-vector}

\paragraph{Definition.}
Task vector arithmetic represents task-induced behavior as a parameter difference between a fine-tuned model and its pretrained counterpart (PLM), and applies linear addition or subtraction to transplant or suppress behaviors in another model \citep{ilharco2022editing}. Its appeal is recombining task knowledge in parameter space without further training.

\paragraph{Design and compatibility.}
Element-wise composition is most stable when models share the same architecture \citep{ilharco2022editing}. LayerNorm parameters are known to be sensitive \citep{xiong2020onlayernorm}. Thus, prior work often excludes them during vector operations \citep{shirafuji2024bias}. When composing multiple vectors, interference and sign conflicts can degrade performance. Therefore, standardization and merging procedures have been proposed to mitigate these effects \citep{yadav2023tiesmerging}.

\paragraph{Applications and limits.}
The idea extends to alignment use cases. \emph{Chat vectors} align dialogue capability by composing a chat–PLM difference into a continual pretraining model (CP Model) in another language, yielding instruction following \citep{huang2023chat}. \emph{Bias vectors} subtract biases learned from curated corpora \citep{shirafuji2024bias}. These approaches often require prepared data for the source behavior and have been shown primarily on small or medium models, leaving generalization to larger LLMs as an open question.

\paragraph{Relation to distillation.}
Distillation is a natural, architecturally flexible alternative for transferring guard behavior across languages, but it addresses a different bottleneck from Guard Vector. Distillation requires target-language inputs, teacher-generated hard labels or logits, and an additional student-training stage, so its effectiveness is bounded by the teacher guard's reliability on the target language and domain. Guard Vector instead transfers guard behavior in parameter space without target-language safety labels or teacher-generated labels, but requires an architecture-compatible target-language CP backbone. Distillation therefore has a supervision/teacher-reliability bottleneck, whereas Guard Vector has a compatible-backbone availability bottleneck; our setting targets the latter, and we do not claim that Guard Vector is universally preferable to distillation.

\subsection{Streaming-Aware Guardrails for Real-World Deployment}
\label{app:streaming}

\paragraph{Gaps in streaming evaluation protocols.}
Recent work enables streaming or long-context generation (e.g., attention sinks, windowed/long-context attention, compressed memories) \citep{xiao2023efficient, han2023lm, munkhdalai2024leave}.
Industry toolkits also expose token- or chunk-level callbacks for checks during generation \citep{nvidia_nemo_guardrails_streaming_guide_2025, nvidia_nemo_guardrails_config_guide_2025}.
However, for guardrail classifiers specifically, standardized streaming evaluation remains under-specified. Few reports define comparable prefix-time protocols, early-termination policies, and shared metrics that relate streaming decisions to offline ground truth (e.g., F1, BER) together with prefix-timing measures such as TTD, or verify parity between streaming and offline results.

\paragraph{Inference cost and user experience in streaming.}
Public guardrails commonly compute per-harm scores and aggregate to a binary outcome, and many implementations rely on multi-token generation at inference (e.g., label descriptions or rationales) \citep{zeng2024shieldgemma, dubey2024llama3herdmodels}.
Under streaming environment, these pipelines must run on each growing prefix. When categories are scored separately, cost scales with the number of harms and extends the decode loop.
This raises tail latency and slows time-to-first-decision under concurrency, which can surface as delayed blocks or visible lag for end users.

\clearpage
\section{Limitations}
\label{app:limitations}

\paragraph{Scope of the training-free claim.}
Our training-free, label-free claim applies specifically to the Guard Vector composition step. The overall pipeline still assumes an aligned source Guard Model, its same-architecture base PLM, and a target-language continual-pretraining (CP) checkpoint; constructing these checkpoints requires training. The streaming-aware prefix SFT stage is a separate supervised specialization that uses curated prefix data.

\paragraph{Architecture and checkpoint requirements.}
Guard Vector composition assumes that the PLM, Guard Model, and CP Model share the same architecture and compatible parameter shapes; composing across different model families is out of scope. Applicability is also limited for languages that lack a suitable CP checkpoint.

\paragraph{Streaming evaluation is Korean-centered.}
Guard Vector composition is evaluated across Chinese, Japanese, and Korean in the offline regime, but our streaming-aware prefix SFT is validated primarily on Korean. Absolute classification performance also varies by language and dataset difficulty. Extending streaming-aware prefix SFT and its evaluation to additional languages remains future work. Chinese/Japanese prefix SFT streaming results and multilingual baselines are reported in Appendix~\ref{app:cjk-streaming}, English source-language retention in Appendix~\ref{app:src-retention}, and adversarial/OOD stress tests in Appendix~\ref{app:adversarial}; the latter remain challenging, especially for Japanese under streaming.

\paragraph{Binary classification scope.}
Our evaluation targets binary SAFE/UNSAFE response classification, matching our streaming-blocking deployment goal. Category-level (multi-class) safety attribution---e.g., a single-token category interface such as \texttt{<SAFE>/<UNSAFE-S1>/\ldots}---is compatible with our prefix SFT format but requires a separate evaluation protocol and is left for future work.

\paragraph{Threshold robustness.}
As shown in Appendix~\ref{subsec:threshold-sensitivity}, streaming decisions can be sensitive to the unsafe classification threshold, especially under realistic SAFE-heavy traffic. While TGM~+~prefix SFT stays stable over a practical high-precision range there (Table~\ref{tab:krai-threshold}), we do not claim robustness at arbitrary thresholds or under distribution shift. We therefore recommend conservative, stability-oriented threshold policies and treating threshold robustness as a first-class deployment requirement.

\end{document}